\documentclass[sigconf]{acmart}
\usepackage{balance}
\usepackage{color}
\usepackage{colortbl}
\usepackage{subfigure} 
\usepackage{amsmath}

\usepackage{amssymb} 
\usepackage{booktabs}
\usepackage{multirow}
\usepackage{microtype}
\usepackage{graphicx}

\usepackage{mathrsfs}
\usepackage{hyperref}
\usepackage{algorithm}%
\usepackage{algorithmic}
\newcommand{\bp}[1]{\textcolor{black}{#1}}

\renewcommand{\shortauthors}{Liu, et al.}
\AtBeginDocument{%
  \providecommand\BibTeX{{%
    \normalfont B\kern-0.5em{\scshape i\kern-0.25em b}\kern-0.8em\TeX}}}

\copyrightyear{2021}
\acmYear{2021}
\setcopyright{acmcopyright}\acmConference[MM '21]{Proceedings of the 29th ACM International Conference on Multimedia}{October 20--24, 2021}{Virtual Event, China}
\acmBooktitle{Proceedings of the 29th ACM International Conference on Multimedia (MM '21), October 20--24, 2021, Virtual Event, China}
\acmPrice{15.00}
\acmDOI{10.1145/3474085.3475474}
\acmISBN{978-1-4503-8651-7/21/10}

\begin{document}
\fancyhead{}
\title{Exploring Gradient Flow Based Saliency for DNN Model Compression}

\author{Xinyu Liu$^{1\ast}$, Baopu Li$^{2\ast}$, Zhen Chen$^{1}$, Yixuan Yuan$^{1\ddagger}$}

\affiliation{
\institution{$^{1}$City University of Hong Kong}
\institution{$^{2}$Baidu USA LLC}
\institution{\{xliu423-c, zchen72-c\}@my.cityu.edu.hk, baopuli@baidu.com}
\institution{\{yxyuan.ee\}@cityu.edu.hk}
\city{}
\country{}
}

\thanks{$\ast$ Equal contribution. $\ddagger$ Corresponding Author.}

\begin{abstract}
Model pruning aims to reduce the deep neural network (DNN) model size or computational overhead. Traditional model pruning methods such as $\ell_1$ pruning that evaluates the channel significance for DNN pay too much attention to the local analysis of each channel and make use of the magnitude of the entire feature while ignoring its relevance to the batch normalization (BN) and ReLU layer after each convolutional operation. To overcome these problems, we propose a new model pruning method from a new perspective of gradient flow in this paper. Specifically, we first theoretically analyze the channel's influence based on Taylor expansion by integrating the effects of BN layer and ReLU activation function. Then, the incorporation of the first-order Talyor polynomial of the scaling parameter and the shifting parameter in the BN layer is suggested to effectively indicate the significance of a channel in a DNN. Comprehensive experiments on both image classification and image denoising tasks demonstrate the superiority of the proposed novel theory and scheme. Code is available at \url{https://github.com/CityU-AIM-Group/GFBS}.
\end{abstract}
	
	\maketitle

\section{Introduction}

The remarkable performance in multimedia tasks brought by deep neural networks (DNNs) is accompanied with  huge computation resource cost  \cite{cai2018proxylessnas}. To embed them in resource-constrained scenarios and reduce their storage, model compression techniques \cite{lecun1989optimal, jaderberg2014speeding, li2016pruning} are proposed to meet the budgets. Among which, channel pruning has shown promising results in overcoming the prohibition of deploying DNNs to mobile or wearable devices. Many  channel pruning methods start with a pre-trained large scale DNN, analyze the importance of each channel (convolutional filter) and then remove the redundant ones under a certain criterion. This strategy intends to maintain the performance of the original network as much as possible and can enjoy the off-the-shelf deep learning libraries for practical acceleration.

Although numerous channel pruning algorithms have been proposed, there remain several open issues. The first is that \textit{most criteria are heuristic}. For example, \citep{li2016pruning} used the smaller $\ell_p$ norm less informative assumption to remove the channels, and \cite{slimming} selected filters to be pruned according to the scaling parameter $\gamma$ in the batch normalization (BN) layer \cite{ioffe2015batch} in the network. Despite the notable pruned ratio and accuracy obtained, discovering which subsets of the channels inside a pre-trained DNN have the largest saliencies is nontrivial. Therefore, we argue that revealing the channel saliencies via rigorous deduction based on certain criteria could be critical in pruning the neural networks.

Along the line of pruning with significance of channels, oracle pruning shows its great advantage since it comprehensively investigates the impact of each channel in a network. To overcome the great computational cost in oracle pruning strategy, Molchannov \textit{et.al.} applied the elegant analysis of Taylor expansion \cite{molchanov2016pruning} to simplify the greedy searching process of orcle pruning.
In addition, the input response of the next layer \cite{luo2017thinet} and a binary search for optimal channels in each layer \cite{aofp} were also suggested to facilitate the one-by-one based channel evaluation process.
However, all the above representative methods either collect all the numerical values in the feature maps or overlook the BN and ReLU activations \cite{nair2010rectified} in modern architectures. \textit{So, generally speaking, these methods may be inefficient and lack a holistic overview of the network structure.}

To address the above issues, we aim to utilize the \textbf{G}radient \textbf{F}low during  \textbf{B}ack propagation for revealling the channel \textbf{S}aliencies (GFBS) within a network and propose a new model pruning method based on this viewpoint, namely GFBS. 
Specifically,  we take advantage of the Taylor expansion as \cite{molchanov2016pruning} did but in a more efficient and holistic manner. In addition, key components such as BN and ReLU that are widely used are also leveraged into the above process. Therefore, our method is considered more thorough and legitimate. Suprisingly, theoretical analysis brings to the conclusion that channel saliencies can be re-formulated with \textit{the combination of the first-order Taylor polynomial of the scaling parameter and the signed shifting parameter in the BN layer.} With GFBS, the channel-wise importance could be captured directly without accessing the values inside the convolutional layers. Based on this insight, we are able to rank the importance of the filters effectively within one forward and backpropagation of a minibatch instead of evaluating all channels in sequence. Besides, the proposed method induces no extra trainable parameters and prior regularizations, thus is friendly for implementation on arbitrary networks and tasks. Extensive image classification and image denoising experiments show the superior performance of the proposed model compression scheme.

In summary, the major contributions of our work are three fold: 

\begin{itemize}
    \vspace{-2pt}
    \item  We theoretically analyze the channel significance problem from a holistic perspective of gradient flow. Such a global view enables more effective description of network structure. It also contains no extra trainable parameters, and can be widely applied to different types of neural networks. 
    
    \item We propose a novel model pruning method GFBS that decides the channel saliencies by combining the first order Taylor polynomial of the scaling parameter and the signed shifting parameter in the BN layer, leading to more effective and efficient channel pruning in a network. 
    
   \item We comprehensively validate the proposed novel scheme on two common tasks image classification and image denoising. Several benchmark datasets with typical models such as VGGNet, ResNet, MobileNetv2 and DenseNet for classification and DnCNN for denoising are tested, our method shows a superior model compression performance.
   \vspace{-2pt}
\end{itemize}

\section{Related Works}

\subsection{Unstructured Pruning}
Trimming the neural networks has been widely studied by researchers since the last century. Unstructured pruning aims at locating and removing connections that have a negligible influence on the prediction accuracy of the network. To find out these connections, the error with the removal of each unit was computed \cite{mozer1989skeletonization}. Through calculating the integral of the error change in the error function during the removal of each weight, the sensitivity of each connection was estimated via backpropagation \cite{karnin1990simple}. Other prior  {authors} also proposed saliency measurement methods for removing redundant weights according to their corresponding Hessian matrix  {of} the loss function \cite{lecun1989optimal, hassibi1992second}. \cite{molchanov2017variational} studied the variational dropout \cite{kingma2015variational} and intended to prune weights with high dropout rates. A recent work \cite{lee2018snip} suggested an approach to update the importance of weights via backpropagating gradients from the loss function. However, \bp{the unstructured pruning will form a sparse matrix that is not friendly to practical implementation. As such, recently more attention has been paid to structured pruning.}

\subsection{Structured Pruning}
Structured pruning, also known as filter pruning, is one of the most widely used techniques for achieving inference acceleration of the neural networks nowadays \cite{li2016pruning, ssl, molchanov2016pruning, luo2017thinet, slimming, sss, tandl, fpgm, zhao2019variational, gordon2018morphnet, molchanov2019importance, hinge, ning2020dsa, kang2020operation, legr, lin2020hrank}. By evaluating the importance of the filters, those that are unlikely to have a great contribution to the accuracy of a network can be safely removed. Instead of pruning the weight matrices, structured pruning imposes channel-wise sparsity on the feature maps thus will inherit the benefits given by the Basic Linear Algebra Subprograms (BLAS) libraries. 

\noindent \textbf{Heuristic Metrics.} Similar to the criterion proposed in unstructured pruning, the channel-wise $\ell_1$ norm criterion was first studied \cite{li2016pruning}. Then, APoZ  {leveraged} another heuristic metric that prunes unimportant filters according to the average percentage of zeros \cite{APoZ}. Network slimming \cite{slimming} added regularization on the BN layers to select channels with larger scaling parameters $\gamma$. FPGM filtered out redundant channels with a criterion of the geometric median \cite{fpgm}. The oracle pruning ranks the filters according to their {impacts} on the final loss of the neural network without that filter.
It is also named as \textit{channel saliency} because it corresponds to the importance of each channel-wise feature. However, it will lead to intolerable time consumption practically because each filter needs to be assessed consecutively. Some methods have attempted to approximate the process via mathematical deduction. \cite{molchanov2016pruning} modelled the importance of each feature map as the first order Taylor expansion with regards to the loss function. ThiNet \cite{luo2017thinet} pruned each layer with the statistics information computed from its next layer. AOFP \cite{aofp} searched for the least important filters with binary search.

\noindent \textbf{Neural Architecture Search and Meta-Learning.} Apart from {explicit} assessment of the importance, a new branch that automatically prunes the network through Neural architecture search (NAS) or Meta-learning emerged recently. \cite{li2020dhp} searched for pruning strategies within a hypernetwork by taking latent vectors as inputs. \cite{ning2020dsa} searched for inter-layer sparsity with budgeted pruning from scratch. \cite{liu2019metapruning} proposed a meta network to generate weights for various pruned structures. 

Despite the superiority in performance, most structured saliency evaluation methods: 1)  are based on heuristic criteria or induce auxiliary trainable parameters or networks, which may be insufficient in theoretical supportance and mathematical model analysis; 2) rely on all numerical values in the features which may lead to intensive computation to determine the unimportant channels; 3) conduct saliency analysis layer-wise, thus lack a holistic overview of the channels throughout the network. Therefore,  we opt to discovering the authentic channel saliencies via gradient flow during backpropagation without extra searching or training, which is in contrast to all the above works in structured pruning.

\section{Methodology}
\subsection{Problem Formulation and Notations}

\textbf{Problem Formulation.} We first take a brief review of modern DNN architectures. Enormous novel deep neural network architectures are proposed with various patterns in recent years. The VGGNet equipped with BN \cite{simonyan2014very, ioffe2015batch} has shown better performance in many vision tasks than the original version. ResNet \cite{he2016deep} introduced residual blocks that contain skip connections to avoid network degeneration. MobileNet series \cite{howard2017mobilenets, sandler2018mobilenetv2, howard2019searching} used depthwise convolutions to reduce computation cost meanwhile retain the orginal network performance. It is observed that even though their designs are based on different intentions, the basic unit of networks shares an intrinsic similarity that they all have a \texttt{Conv-BN-ReLU} layout\footnote{An exception lies in the MobileNetv2 architecture, the convolutional layers before the shortcut addition remove the ReLU activation function, which can form a linear bottleneck to capture the low-dimensional manifold of interest.}. However, there lack pruning works that focus on this aspect, and we conjecture that previous literature that utilizes the magnitudes in convolutional layers or scaling parameters in BN layers independently could not cover all the embedded information that may be beneficial to evaluate saliency of each channel in DNNs.

\noindent \textbf{Notations.} For a convolutional neural network $\Theta$ with $L$ layers that is trained on a dataset $\mathcal{D}$, the output feature map from each layer is represented by $\mathcal{F}_l$ $\in$ $\mathbb{R}^{\mathcal{H}_l \times \mathcal{W}_l \times \mathcal{C}_l}$, where $ \mathcal{C}_l$ represents the channel dimension and $l \in \{1, ..., L\}$. Therefore, $\mathcal{F}_l^{(j)}$ denotes the channel with an index $j$ in the $l$-th layer, where $j \in \{1, ..., \mathcal{C}_l\}$. For a convolutional layer
with \bp{a kernel size of} $k_l \times k_l$, it comprises a weight tensor $\mathbf{W}_l$ with size $k_l \times k_l \times \mathcal{C}_{l-1} \times \mathcal{C}_l$ and a bias tensor $\mathbf{b}_l$ with size $\mathcal{C}_l$. The following BN layer also contains two learnable affine parameters, namely $\gamma_l$ and $\beta_l$, 
both have a shape of $\mathcal{C}_l$. Accordingly, the overall parameters in layer $l$ are denoted as $\mathbf{P}_l = \{\mathcal{D}|(\mathbf{W}_l, \mathbf{b}_l, \gamma_l, \beta_l)\}$ and the parameters for generating $\mathcal{F}_l^{(j)}$ is $\mathbf{P}_l^{(j)} = \{\mathcal{D}|(\mathbf{W}_l^{(j)}, \mathbf{b}_l^{(j)}, \gamma_l^{(j)}, \beta_l^{(j)})\}$. The training minibatch $\mathcal{B}$ is composed of $m$ samples.

\subsection{Gradient Flow Based Saliency}
A straightforward measurement of the channel saliencies is the error fluctuation when a channel-wise filter $\mathcal{F}_l^{(j)}$ in a pre-trained network is pruned. To approximate this term, we trace the forward and backpropagation process of generation and  update of the relevant parameters $\mathbf{P}_l^{(j)}$. Specifically, we first consider the authentic channel saliency that corresponds to the error fluctuation when each channel-wise filter $\mathcal{F}_l^{(j)}$ in a pre-trained network is pruned, then take BN and ReLU into account sequentially.

\noindent \textbf{Lemma 3.1.} For a convolutional layer, 
removing its channel-wise output feature $\stackrel{\sim}{\mathcal{F}}_l^{(j)}$ is equivalent to allocate $\mathbf{W}_l^{(j)}=\mathbf{b}_l^{(j)}=0$.

\noindent \textit{Proof.} Given a convolutional layer with weight $\mathbf{W}_l^{(j)}$ and bias $\mathbf{b}_l^{(j)}$, the channel-wise feature ${\stackrel{\sim}{\mathcal{F}}_l^{(j)}}$ 
can be produced by
\vspace{-4pt}
\begin{equation}
    {\stackrel{\sim}{\mathcal{F}}_l^{(j)}}=\mathbf{W}_l^{(j)}\times\mathcal{F}_{l-1}+\mathbf{b}_l^{(j)},
    \vspace{-4pt}
\label{2}
\end{equation}
where $\mathcal{F}_{l-1}$ is output feature of the last layer, which is considered as a constant if the input data is fixed. Then, $\mathbf{W}_l^{(j)}=\mathbf{b}_l^{(j)}=0$ will make ${\stackrel{\sim}{\mathcal{F}}_l^{(j)}}=0$.

\noindent \textbf{Lemma 3.2.} The \textit{channel saliencies} of the feature map after the BN layer can be approximated by the first-order Taylor polynomial of the scaling parameter $\gamma_l^{(j)}$.

\noindent \textit{Proof.} 
Based on the measurement of the relevance of a unit in \cite{mozer1989skeletonization}, the authentic saliencies of each channel can be defined as:
\vspace{-2pt}
\begin{equation}
    |\mathcal{S}(\stackrel{\sim}{\mathcal{F}}_l^{(j)})|=|\mathcal{L}({\mathcal{D}|\Theta(\stackrel{\sim}{\mathcal{F}}_l^{(j)}=0)})-\mathcal{L}({\mathcal{D}|\Theta})|,
    \vspace{-4pt}
\end{equation}
which corresponds to assessing the error fluctuation with the removal of each channel in the network consecutively. Then, by taking advantage of the Taylor expansion, we expand the loss function at $\stackrel{\sim}{\mathcal{F}}_l^{(j)}=0$:
\vspace{-8pt}
\begin{equation}
\begin{split}
    \mathcal{L}({\mathcal{D}|\Theta(\stackrel{\sim}{\mathcal{F}}_l^{(j)}=0)})=\mathcal{L}({\mathcal{D}|\Theta})-J(\stackrel{\sim}{\mathcal{F}}_l^{(j)})\stackrel{\sim}{\mathcal{F}}_l^{(j)}+R_1,
\end{split}
\end{equation}
\noindent where $R_1$ is the higher order Lagrange remainder and $J(\stackrel{\sim}{\mathcal{F}}_l^{(j)})$ is the Jacobian matrix
\begin{equation}
    J(\stackrel{\sim}{\mathcal{F}}_l^{(j)}) = \frac{\partial\mathcal{L}}{\partial\stackrel{\sim}{\mathcal{F}}_l^{(j)}}.
\end{equation}
Considering the computational complexity and the small absolute value of the Lagrange remainder, we omit it and substitute $\mathcal{S}(\stackrel{\sim}{\mathcal{F}}_l^{(j)})$ in:
\begin{equation}
\begin{aligned}
    \mathcal{S}(\stackrel{\sim}{\mathcal{F}}_l^{(j)}) &= |\mathcal{L}({\mathcal{D}|\Theta})- \frac{\partial\mathcal{L}}{\partial\stackrel{\sim}{\mathcal{F}}_l^{(j)}}\stackrel{\sim}{\mathcal{F}}_l^{(j)}-\mathcal{L}({\mathcal{D}|\Theta})|\\
    &=|\frac{\partial\mathcal{L}}{\partial\stackrel{\sim}{\mathcal{F}}_l^{(j)}}\stackrel{\sim}{\mathcal{F}}_l^{(j)}|.
\end{aligned}
\end{equation}
Based on Lemma 3.1 and the Taylor expansion of two variables, we have
\vspace{-8pt}
\begin{equation}
\begin{aligned}
    |\mathcal{L}&({\mathcal{D}|\Theta(\stackrel{\sim}{\mathcal{F}}_l^{(j)}=0)})|= |\mathcal{L}({\mathcal{D}|\Theta(\mathbf{W}_l^{(j)}}=\mathbf{b}_l^{(j)}=0))|,
\end{aligned}
\end{equation}
\begin{equation}
    \mathcal{S}({\stackrel{\sim}{\mathcal{F}}_l^{(j)}}) = |J(\mathbf{W}_l^{(j)}, \mathbf{b}_l^{(j)})(\mathbf{W}_l^{(j)}, \mathbf{b}_l^{(j)})|,
\label{saliencyformula}
\end{equation}
such that the channel saliency after the convolutional layer is equivalent to the first-order Taylor polynomial of the parameters in the convolutional filters. Then, we recall the BN computation:
\begin{equation}
    \mu_l^{(j)} = \frac{1}{\mathcal{C}_l}\sum_{i=1}^{\mathcal{C}_l}\stackrel{\sim}{\mathcal{F}}_{l,i}^{(j)}, 
    \enspace
    (\sigma_l^{(j)})^2 = \frac{1}{\mathcal{C}_l}\sum_{i=1}^{\mathcal{C}_l}(\stackrel{\sim}{\mathcal{F}}_{l,i}^{(j)}-\mu_l^{(j)})^2,
\label{forwardbn1}
\end{equation}
\begin{equation}
    \stackrel{\land}{\mathcal{F}}_l^{(j)} = \frac{\stackrel{\sim}{\mathcal{F}}_l^{(j)} - \mu_l^{(j)}}{\sqrt{(\sigma_l^{(j)})^2 + \epsilon}}, 
    \enspace
    \overline{\mathcal{F}_l^{(j)}} =\gamma_l^{(j)}\stackrel{\land}{\mathcal{F}}_l^{(j)} + \beta_l^{(j)},
\label{forwardbn2}
\end{equation}
where $\epsilon$ is a small positive value to avoid dividing by zero. Hence, we have:
\vspace{-6pt}
\begin{equation}
\begin{aligned}
    \overline{\mathcal{F}_l^{(j)}}
    =\gamma_l^{(j)}\frac{\stackrel{\sim}{\mathcal{F}}_l^{(j)} - \frac{1}{\mathcal{C}_l}\sum_{i=1}^{\mathcal{C}_l}\stackrel{\sim}{\mathcal{F}}_{l,i}^{(j)}}{\sqrt{\frac{1}{\mathcal{C}_l}\sum_{i=1}^{\mathcal{C}_l}(\stackrel{\sim}{\mathcal{F}}_{l,i}^{(j)}-\mu_l^{(j)})^2 + \epsilon}} + \beta_l^{(j)}.
\end{aligned}
\label{gamma1}
\end{equation}
From Equation (\ref{2}), (\ref{forwardbn1}), (\ref{forwardbn2}), which are the forward of a BN layer, 
\vspace{-6pt}
\begin{equation}
\begin{aligned}
    &\overline{\mathcal{F}_l^{(j)}} =\gamma_l^{(j)}\frac{(\mathbf{W}_l^{(j)}\times\mathcal{F}_{l-1}+\mathbf{b}_l^{(j)})}{\sqrt{\frac{1}{\mathcal{C}_l}\sum_{i=1}^{\mathcal{C}_l}(\mathbf{W}_{l,i}^{(j)}\times\mathcal{F}_{l-1}+\mathbf{b}_{l,i}^{(j)})^2 + \epsilon}} \\
    &- \gamma_l^{(j)}\frac{\frac{1}{\mathcal{C}_l}\sum_{i=1}^{\mathcal{C}_l}(\mathbf{W}_{l,i}^{(j)}\times\mathcal{F}_{l-1}+\mathbf{b}_{l,i}^{(j)})}{\sqrt{\frac{1}{\mathcal{C}_l}\sum_{i=1}^{\mathcal{C}_l}(\mathbf{W}_{l,i}^{(j)}\times\mathcal{F}_{l-1}+\mathbf{b}_{l,i}^{(j)})^2 + \epsilon}} + \beta_l^{(j)}.
\end{aligned}
\label{5}
\end{equation}
From Lemma 3.1, for a output feature of a convolutional layer, removing its channel-wise output feature $\stackrel{\sim}{\mathcal{F}}_l^{(j)}$ is equivalent to allocate $\mathbf{W}_l^{(j)}=\mathbf{b}_l^{(j)}=0$, thus Equation (\ref{5}) becomes:
\begin{equation}
\begin{aligned}
    &\overline{\mathcal{F}_l^{(j)}} =\gamma_l^{(j)}\frac{(0\times\mathcal{F}_{l-1}+0)}{\sqrt{\frac{1}{\mathcal{C}_l}\sum_{i=1}^{\mathcal{C}_l}(0\times\mathcal{F}_{l-1}+0)^2 + \epsilon}} \\
    &- \gamma_l^{(j)}\frac{\frac{1}{\mathcal{C}_l}\sum_{i=1}^{\mathcal{C}_l}(0\times\mathcal{F}_{l-1}+0)}{\sqrt{\frac{1}{\mathcal{C}_l}\sum_{i=1}^{\mathcal{C}_l}(0\times\mathcal{F}_{l-1}+0)^2 + \epsilon}} + \beta_l^{(j)},
\end{aligned}
\label{6}
\end{equation}
thus
\begin{equation}
    \overline{\mathcal{F}_l^{(j)}} =\beta_l^{(j)},
\label{conv_only}
\end{equation}
which indicates that for the output feature of the BN layer, removing a channel-wise output feature is equivalent to allocate $\overline{\mathcal{F}_l^{(j)}} =\beta_l^{(j)}$. Then, if we change $\gamma_l^{(j)}=0$ in Equation (\ref{5}):
\begin{equation}
\begin{aligned}
    &\overline{\mathcal{F}_l^{(j)}} =0\times\frac{(\mathbf{W}_l^{(j)}\times\mathcal{F}_{l-1}+\mathbf{b}_l^{(j)})}{\sqrt{\frac{1}{\mathcal{C}_l}\sum_{i=1}^{\mathcal{C}_l}(\mathbf{W}_{l,i}^{(j)}\times\mathcal{F}_{l-1}+\mathbf{b}_{l,i}^{(j)})^2 + \epsilon}} \\
    &- 0\times\frac{\frac{1}{\mathcal{C}_l}\sum_{i=1}^{\mathcal{C}_l}(\mathbf{W}_{l,i}^{(j)}\times\mathcal{F}_{l-1}+\mathbf{b}_{l,i}^{(j)})}{\sqrt{\frac{1}{\mathcal{C}_l}\sum_{i=1}^{\mathcal{C}_l}(\mathbf{W}_{l,i}^{(j)}\times\mathcal{F}_{l-1}+\mathbf{b}_{l,i}^{(j)})^2 + \epsilon}} + \beta_l^{(j)},\\
\end{aligned}
\label{8}
\end{equation}
can also lead to
\begin{equation}
    \overline{\mathcal{F}_l^{(j)}} =\beta_l^{(j)}.
\label{gamma_only}
\end{equation}
which is equivalent to Equation (\ref{conv_only}). Therefore, $\gamma_l^{(j)}=0$ will have the same effect as $\mathbf{W}_l^{(j)}=\mathbf{b}_l^{(j)}=0$ for a \texttt{Conv-BN} block, \textit{thus we can formulate the functional mapping from $\mathbf{W}_l^{(j)}$ and $\mathbf{b}_l^{(j)}$ to $\gamma_l^{(j)}$ as $g_1$}:
\begin{equation}
    \gamma_l^{(j)}=g_1(\mathbf{W}_l^{(j)}, \mathbf{b}_l^{(j)}).
\label{forward}
\end{equation}
Similarly, the first-order derivatives can be computed via tracing the backpropagation of the above process.
After computing the loss of the network $\mathcal{L}$, we have
\begin{equation}
    \frac{\partial\mathcal{L}}{\partial\stackrel{\land}{\mathcal{F}}_{l}^{(j)}} = \frac{\partial\mathcal{L}}{\partial\overline{\mathcal{F}_{l}^{(j)}}}\gamma_l^{(j)},
    \enspace
    \frac{\partial\mathcal{L}}{\partial\gamma_l^{(j)}} = \sum_{i=1}^{\mathcal{C}_l}\frac{\partial\mathcal{L}}{\partial\overline{\mathcal{F}_{l,i}^{(j)}}}\stackrel{\land}{\mathcal{F}}_l^{(j)}.
\label{11_}
\end{equation}
According to the chain rule, for convolutional layers, we have
\begin{equation}
    \frac{\partial\mathcal{L}}{\partial\mathbf{W}_l^{(j)}} = \frac{\partial\mathcal{L}}{\partial\stackrel{\sim}{\mathcal{F}}_{l}^{(j)}}\frac{\partial\stackrel{\sim}{\mathcal{F}}_{l}^{(j)}}{\partial\mathbf{W}_l^{(j)}},
    \enspace
    \frac{\partial\mathcal{L}}{\partial\mathbf{b}_l^{(j)}} = \frac{\partial\mathcal{L}}{\partial\stackrel{\sim}{\mathcal{F}}_{l}^{(j)}}\frac{\partial\stackrel{\sim}{\mathcal{F}}_{l}^{(j)}}{\partial\mathbf{b}_l^{(j)}},
\end{equation}
thus allocating $\frac{\partial\mathcal{L}}{\partial\mathbf{W}_l^{(j)}}$ and $\frac{\partial\mathcal{L}}{\partial\mathbf{b}_l^{(j)}}$ as zero is equal to allocating $\frac{\partial\mathcal{L}}{\partial\stackrel{\sim}{\mathcal{F}}_{l}^{(j)}}$ as zero. Meanwhile, for $\frac{\partial\mathcal{L}}{\partial\stackrel{\sim}{\mathcal{F}}_{l}^{(j)}}$ we have
\begin{equation}
\begin{aligned}
    \frac{\partial\mathcal{L}}{\partial\stackrel{\sim}{\mathcal{F}}_{l}^{(j)}} &= \frac{\partial\mathcal{L}}{\partial\stackrel{\land}{\mathcal{F}}_{l}^{(j)}}\frac{1}{\sqrt{(\sigma_l^{(j)})^2 + \epsilon}} \\
    &+ \frac{\partial\mathcal{L}}{\partial(\sigma_l^{(j)})^2}\frac{2(\stackrel{\land}{\mathcal{F}}_{l}^{(j)}-\mu_l^{(j)})}{m} + \frac{\partial\mathcal{L}}{\partial\mu_l^{(j)}}\frac{1}{\mathcal{C}_l},
\end{aligned}
\label{totalgrad}
\end{equation}
where 
\begin{equation}
    \frac{\partial\mathcal{L}}{\partial(\sigma_l^{(j)})^2} = \sum_{i=1}^{\mathcal{C}_l}\frac{\partial\mathcal{L}}{\partial\stackrel{\land}{\mathcal{F}}_{l,i}^{(j)}}(\stackrel{\land}{\mathcal{F}}_{l,i}^{(j)}-\mu_l^{(j)})\frac{-1}{2}((\sigma_l^{(j)})^2 + \epsilon)^{-3/2},
\label{sigmagrad}
\end{equation}
\begin{equation}
\begin{aligned}
    \frac{\partial\mathcal{L}}{\partial\mu_l^{(j)}} &= (\sum_{i=1}^{\mathcal{C}_l}\frac{\partial\mathcal{L}}{\partial\stackrel{\land}{\mathcal{F}}_{l,i}^{(j)}}\frac{-1}{\sqrt{(\sigma_l^{(j)})^2 + \epsilon}}) \\
    & + \frac{\partial\mathcal{L}}{\partial(\sigma_l^{(j)})^2}\frac{\sum_{i=1}^{\mathcal{C}_l}-2(\stackrel{\sim}{\mathcal{F}}_{l}^{(j)}-\mu_l^{(j)})}{m}.
\end{aligned}
\label{mugrad}
\end{equation}
From Equation (\ref{sigmagrad}) and (\ref{mugrad}), if $\frac{\partial\mathcal{L}}{\partial\stackrel{\sim}{\mathcal{F}}_{l}^{(j)}}$ is zero, both $\frac{\partial\mathcal{L}}{\partial(\sigma_l^{(j)})^2}$ and $\frac{\partial\mathcal{L}}{\partial\mu_l^{(j)}}$ will be zero. Consequently, $\frac{\partial\mathcal{L}}{\partial\stackrel{\sim}{\mathcal{F}}_{l}^{(j)}}$ will equal to zero with regard to Equation (\ref{totalgrad}). Substitute the $\frac{\partial\mathcal{L}}{\partial\stackrel{\sim}{\mathcal{F}}_{l}^{(j)}}$ in Equation (\ref{11_}), $\frac{\partial\mathcal{L}}{\partial\overline{\mathcal{F}_{l}^{(j)}}}$ would be zero since $\gamma_l^{(j)}$ is a fixed value here in backward. Finally, using Equation (\ref{11_}) yields $\frac{\partial\mathcal{L}}{\partial\gamma_l^{(j)}}=0$. 
Therefore, it is implied that the saliency represented by $\frac{\partial\mathcal{L}}{\partial\gamma_l^{(j)}}=0$ is equivalent to the case of $\frac{\partial\mathcal{L}}{\partial\mathbf{W}_l^{(j)}}=\frac{\partial\mathcal{L}}{\partial\mathbf{b}_l^{(j)}}=0$ during backpropagation. Then we can formulate the process with another mapping function $g_2$:
\begin{equation}
    J(\gamma_l^{(j)})=g_2( J(\mathbf{W}_l^{(j)}), J(\mathbf{b}_l^{(j)})).
\label{firstorder}
\end{equation}
\textit{Combining Equation. (\ref{saliencyformula}), (\ref{forward}), and (\ref{firstorder}) yields}:
\begin{equation}
    \mathcal{S}(\overline{{\mathcal{F}_l^{(j)}}}) =  |g_2^{(-1)}(J(\gamma_l^{(j)})) g_1^{(-1)}(\gamma_l^{(j)})|,
\label{afterbn}
\end{equation}
where the right part corresponds to the first-order Taylor polynomial of the scaling parameter $\gamma_l^{(j)}$. 
Note that the mappings above only indicate that the saliencies on both sides of the equation (\ref{forward}), (\ref{firstorder}), and (\ref{afterbn}) are equivalent
. This brings to the conclusion that the channel saliencies of the feature map after the BN layer can be approximated by the first-order Taylor polynomial of $\gamma_l^{(j)}$.

However, as elaborated above, the fragments in the network structures tend to have a \texttt{Conv-BN-ReLU} layout, thus $\beta_l^{(j)}$ that controls which part of the input feature is activated after ReLU also contributes to the discovering of the channel saliencies of the feature map. We theoretically prove this point in the following Theorem 3.3 and evidently validate this statement in the later ablation study.

\noindent \textbf{Theorem 3.3.} For a deep neural network with a basic building block \texttt{Conv-BN-ReLU}, its channel saliencies can be represented by the combination of the first-order Taylor polynomial of $\gamma_l^{(j)}$ and the signed value of $\beta_l^{(j)}$.

\noindent \textit{Proof.} For a channel-wise input feature $\overline{\mathcal{F}_l^{(j)}}$, its corresponding output feature after a ReLU layer is 
\begin{equation}
    \mathcal{F}_l^{(j)} = \rm{max}(\overline{\mathcal{F}_l^{(j)}}, 0).
\end{equation}
Therefore, the partition that gets activated is determined by the positive percentage of $\overline{\mathcal{F}_l^{(j)}}$.
From Equation (\ref{forwardbn1}), (\ref{forwardbn2}), for any arbitrary input,
$\stackrel{\land}{\mathcal{F}}_l^{(j)}$ satisfies the $\mathcal{N}~(0,1)$ normal distribution. After multiplying the scaling parameter, the mean of the distribution remains zero: 
\begin{equation}
    \frac{1}{\mathcal{C}_l}\sum_{i=1}^{\mathcal{C}_l}\gamma_{l,i}^{(j)}\stackrel{\land}{\mathcal{F}}_{l,i}^{(j)} = 0,
\end{equation}
which indicates that for all layers, it is $\beta_l^{(j)}$ that controls the activated percentage uniquely since ReLU only outputs non-zero values for positive inputs, as also suggested by Equation (\ref{conv_only}) and (\ref{gamma_only}). So for each output feature after a ReLU, we use a mapping $g_3$ as the saliency representation:
\begin{equation}
    \overline{\mathcal{S}({\mathcal{F}_l^{(j)}})} = g_3(\mathcal{S}(\overline{{\mathcal{F}_l^{(j)}}}), \beta_l^{(j)}).
\label{afterrelu}
\end{equation}
By jointly considering Equation (\ref{afterbn}) and (\ref{afterrelu}), for the output feature of a \texttt{Conv-BN-ReLU} block, \textit{its channel saliency can be represented using a weighted linear combination as follows:}
\begin{equation}
    \mathcal{S}(\mathcal{F}_l^{(j)}) = |J(\gamma_l^{(j)}) \gamma_l^{(j)}| + \lambda \beta_l^{(j)},
\label{19}
\end{equation}
where $\lambda$ is a balancing term used for achieving the trade-off between the two parts in the criteria, which suggests the proposed theorem. \textit{The deducted Equation (\ref{19}) simply takes advantage of the intermediate information in BN and ReLU to represent the channel saliencies.}
A formal description of GFBS is given in Algorithm \ref{GFBSAlgorithm}.

\begin{algorithm}[t]
   \caption{The GFBS Algorithm}
   \label{GFBSAlgorithm}
\begin{algorithmic}
   \STATE {\bfseries Input:} Training dataset $\mathcal{D}$, Pretrained model ${\Theta}$ = $\{\mathcal{F}_l\}_{l=1}^L$, Balancing term $\lambda$, Desired pruned ratio $\tau$
   \STATE {\bfseries Output:} Pruned network $\Theta’$ = $\{\mathcal{F}_l\}_{l=1}^L$ 
   \STATE {\bfseries Do} Shuffle $\mathcal{D}$ and sample a minibatch $\mathcal{B}$.
   Forward and backpropagate ${\Theta}$ with $\mathcal{B}$ \\
   \STATE Initialize global saliency record set $\mathbf{S}$ $\longleftarrow$ \{\}
   \FOR{$l=1$ {\bfseries to} $L$}
   \FOR{$j=1$ {\bfseries to} $\mathcal{C}_l$}
   \STATE Record $\gamma_l^{(j)}$, $J(\gamma_l^{(j)})$, $\beta_l^{(j)}$ for each $\mathcal{F}_l^{(j)}$
   \ENDFOR \\
   \STATE $\ell_2$ normalize $\gamma_l^{(j)}$, $J(\gamma_l^{(j)})$, $\beta_l^{(j)}$ within each $l$ \\
   \STATE Compute $\mathcal{S}(\mathcal{F}_l^{(j)}) = |J(\gamma_l^{(j)}) \gamma_l^{(j)}| + \lambda \beta_l^{(j)}$ \\
   \STATE Save saliencies $\mathbf{S}$ $\longleftarrow$ $\mathcal{S}(\mathcal{F}_l^{(j)})$
   \ENDFOR \\
   \STATE {\bfseries Do} Sort $\mathbf{S}$, 
  prune the channels in $\mathbf{S}$ until selected channels/total number of channels reach $\tau$ and finetune the network\\
   \STATE {\bfseries Return} Pruned network $\Theta’$ = $\{\mathcal{F}_l\}_{l=1}^L$
   \end{algorithmic}
\end{algorithm}
\begin{figure*}[htbp]
\centering
\subfigure[$\gamma$ distribution in $c$-8.]{
\includegraphics[width=3.4cm]{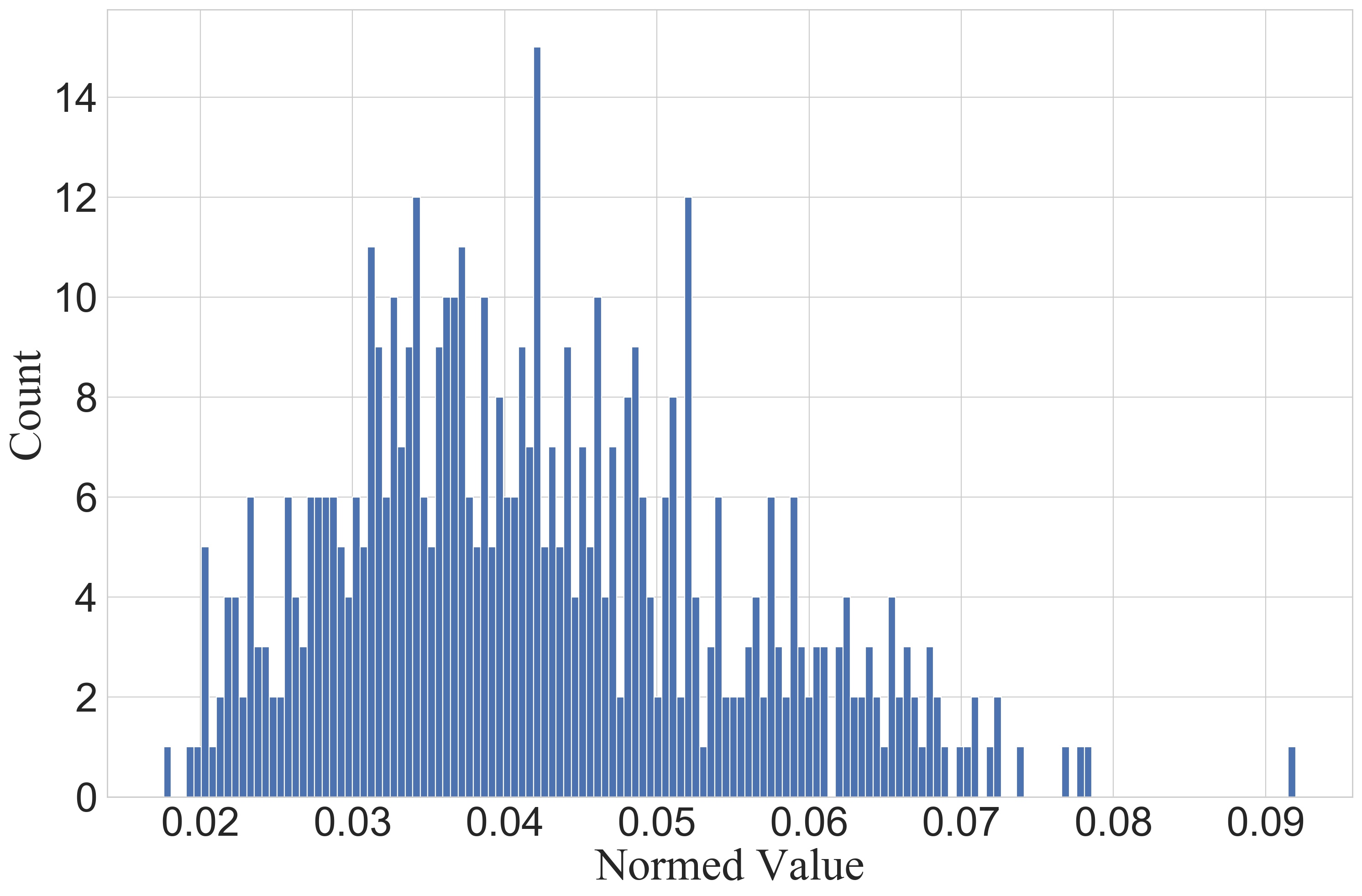}
}
\quad
\subfigure[$\gamma$ distribution in $c$-9.]{
\includegraphics[width=3.4cm]{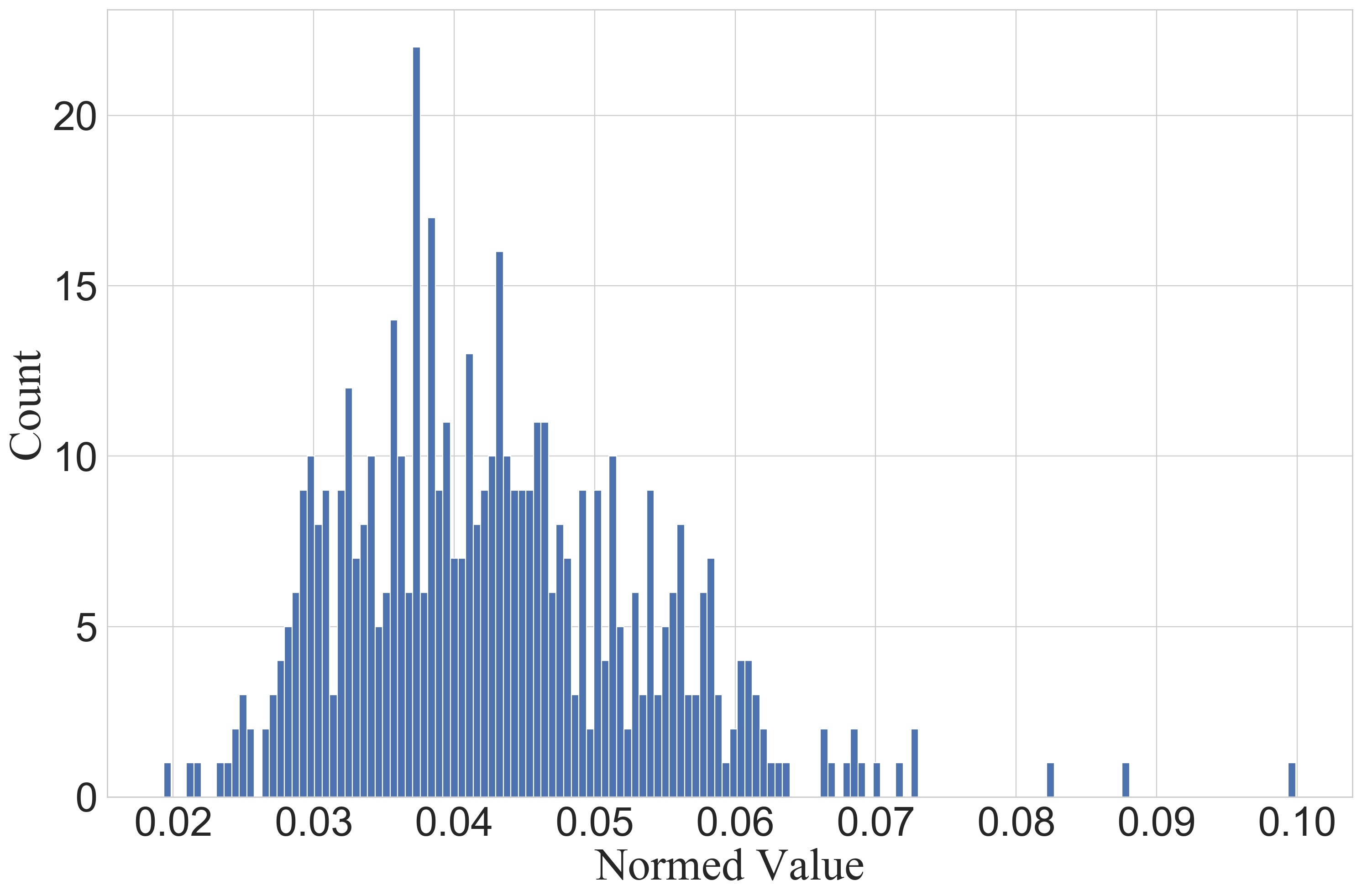}
}
\quad
\subfigure[$\gamma$ distribution in $c$-10.]{
\includegraphics[width=3.4cm]{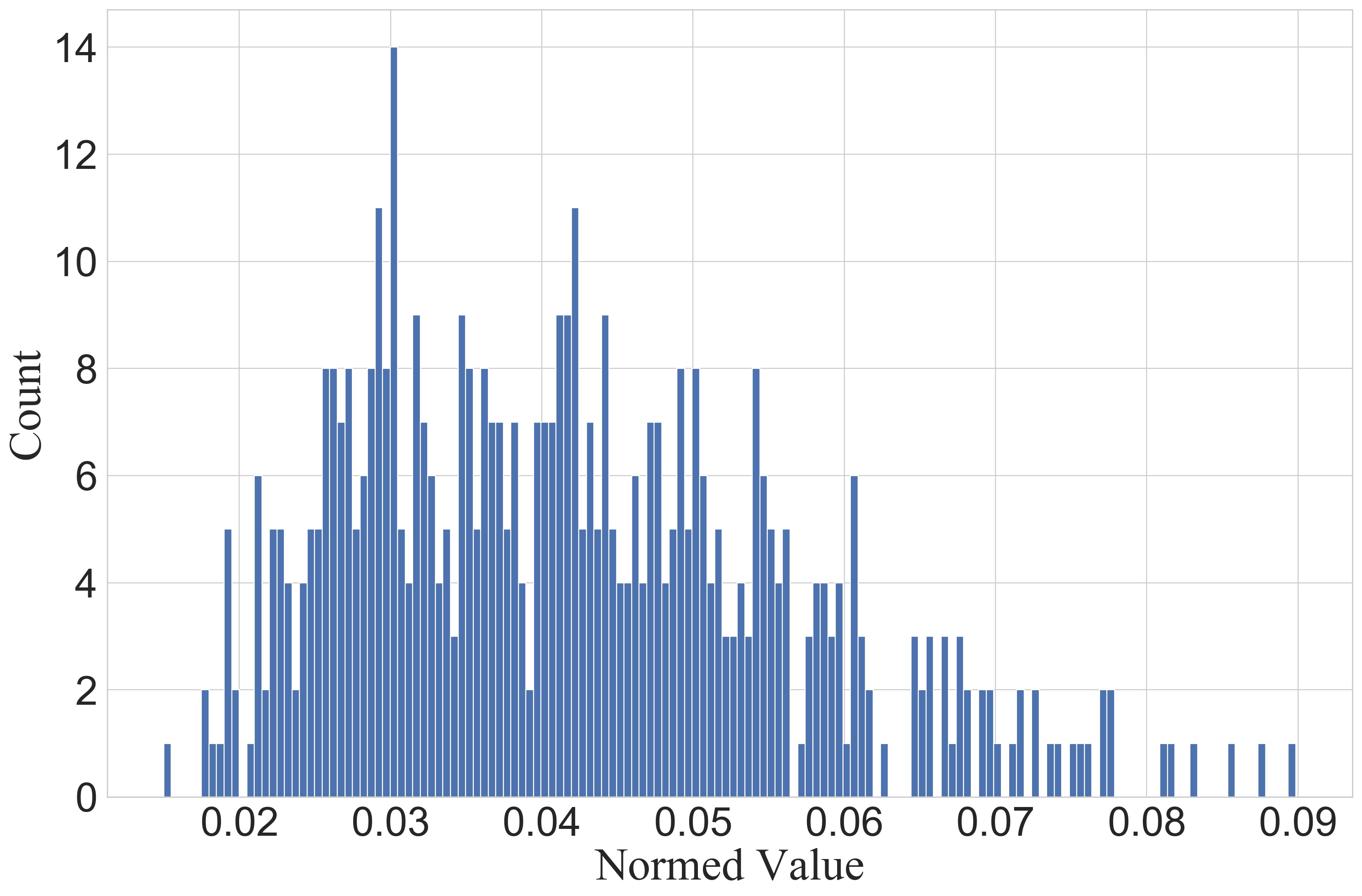}
}
\quad
\subfigure[$\gamma$ distribution in $c$-11.]{
\includegraphics[width=3.4cm]{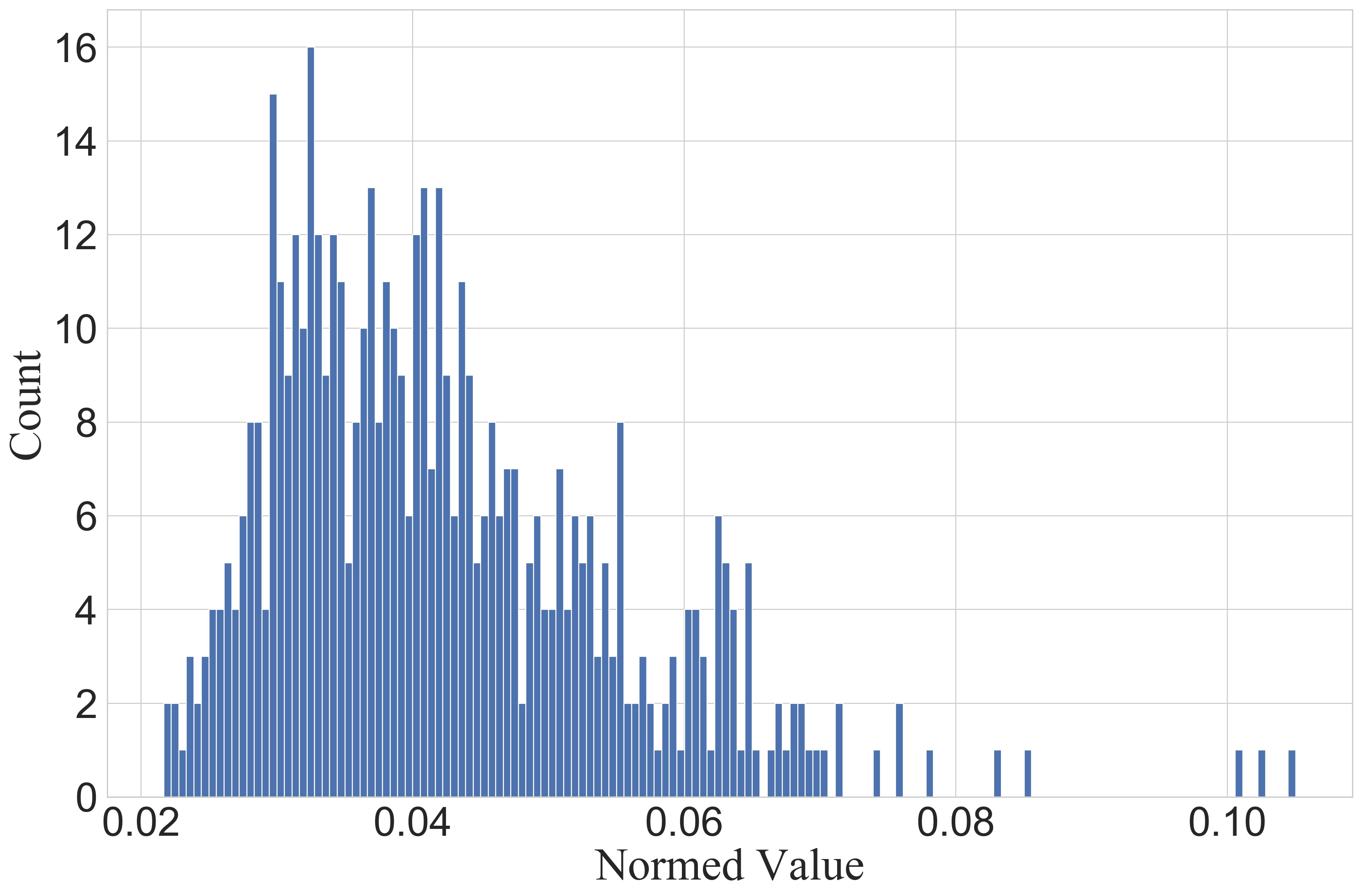}
}
\quad
\subfigure[$\beta$ distribution in $c$-8.]{
\includegraphics[width=3.4cm]{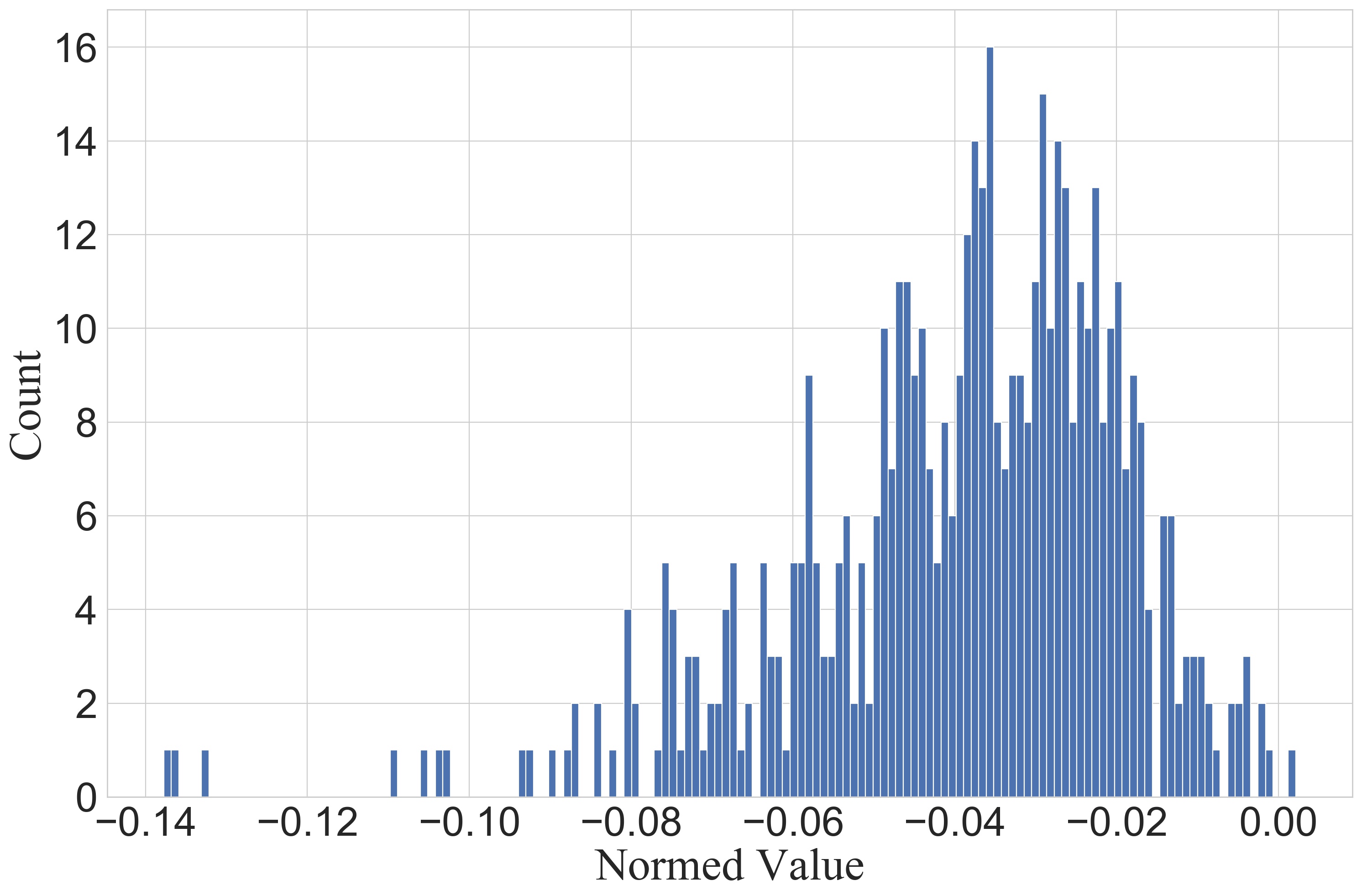}
}
\quad
\subfigure[$\beta$ distribution in $c$-9.]{
\includegraphics[width=3.4cm]{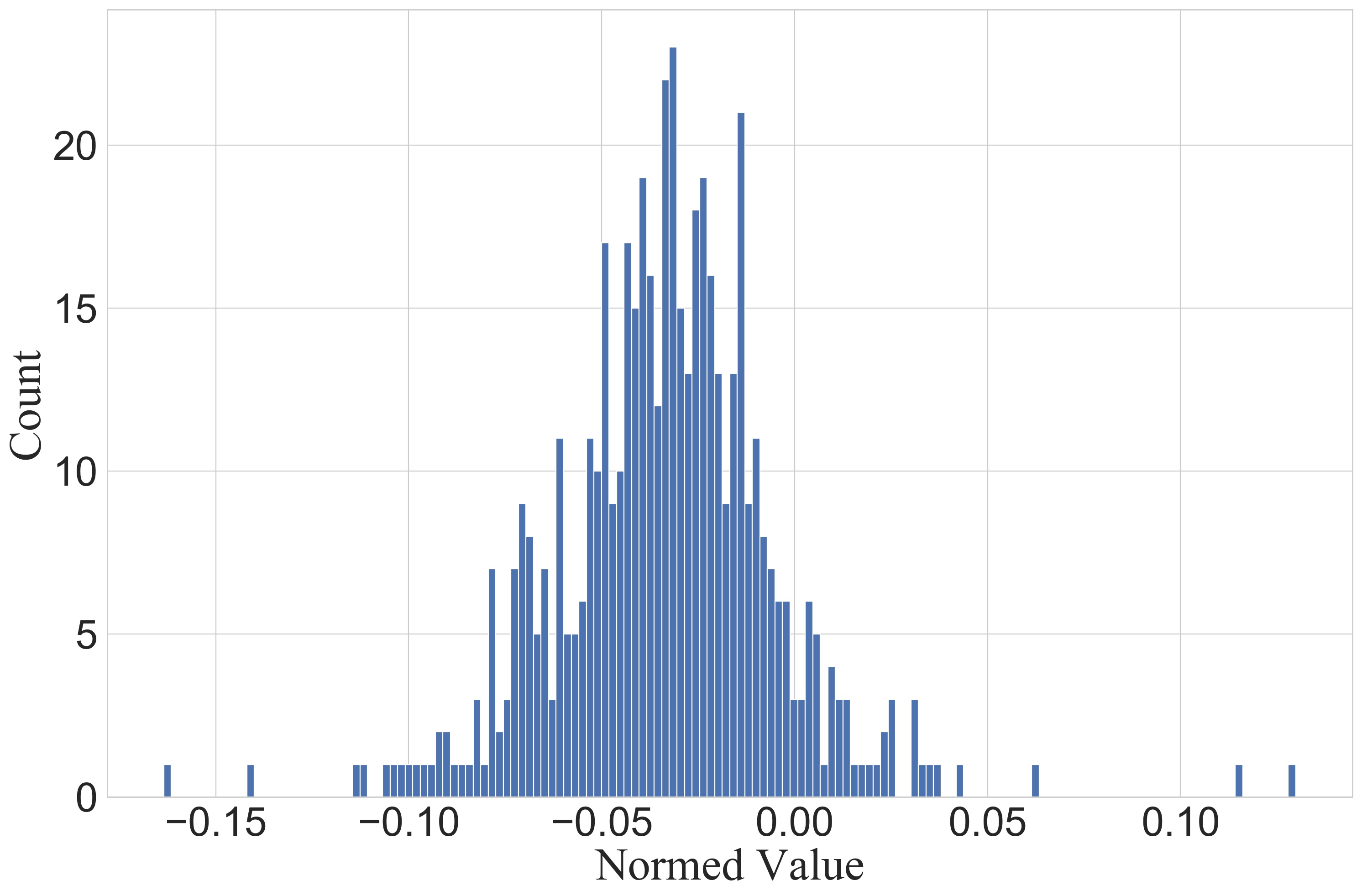}
}
\quad
\subfigure[$\beta$ distribution in $c$-10.]{
\includegraphics[width=3.4cm]{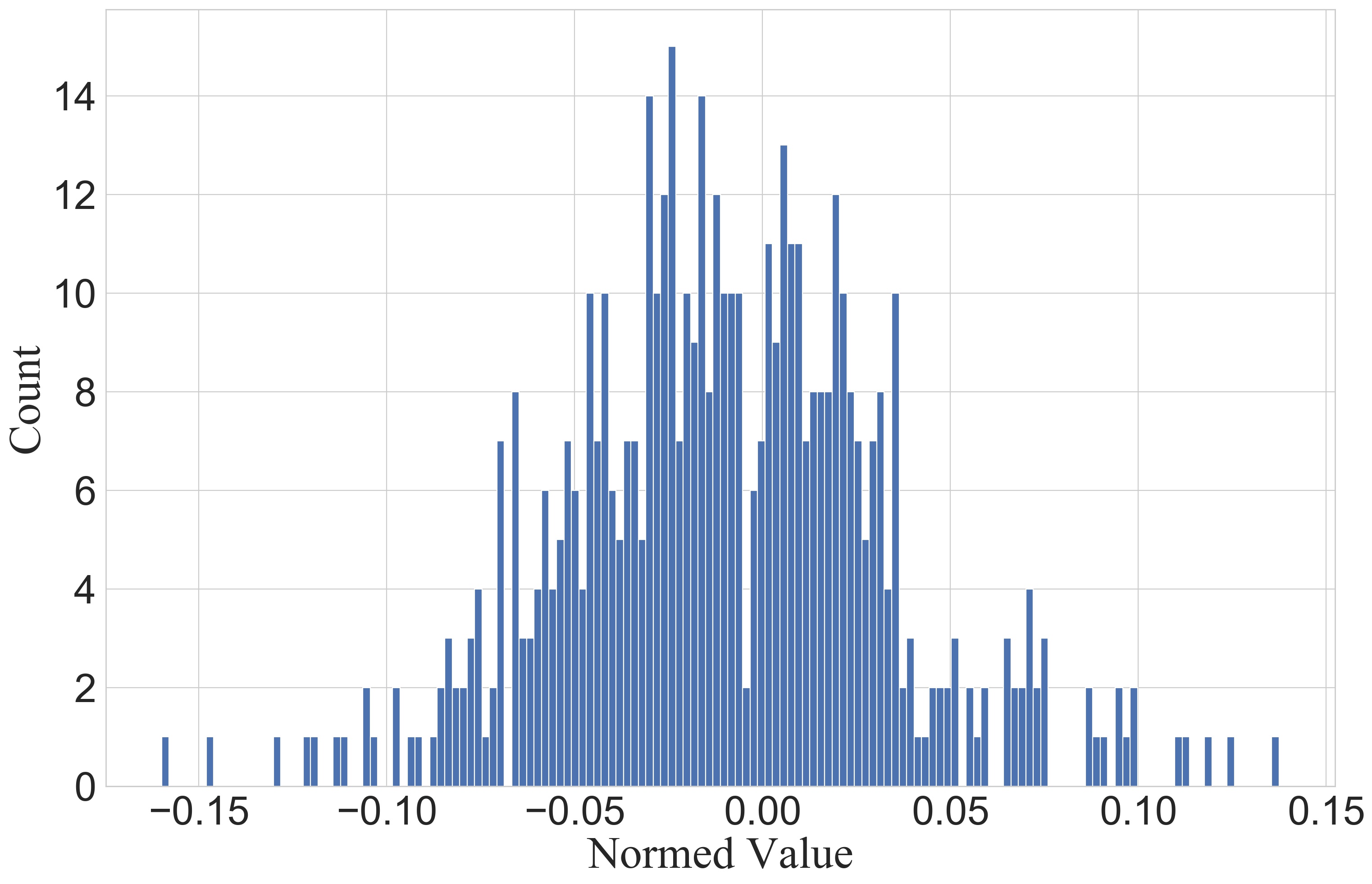}
}
\quad
\subfigure[$\beta$ distribution in $c$-11.]{
\includegraphics[width=3.4cm]{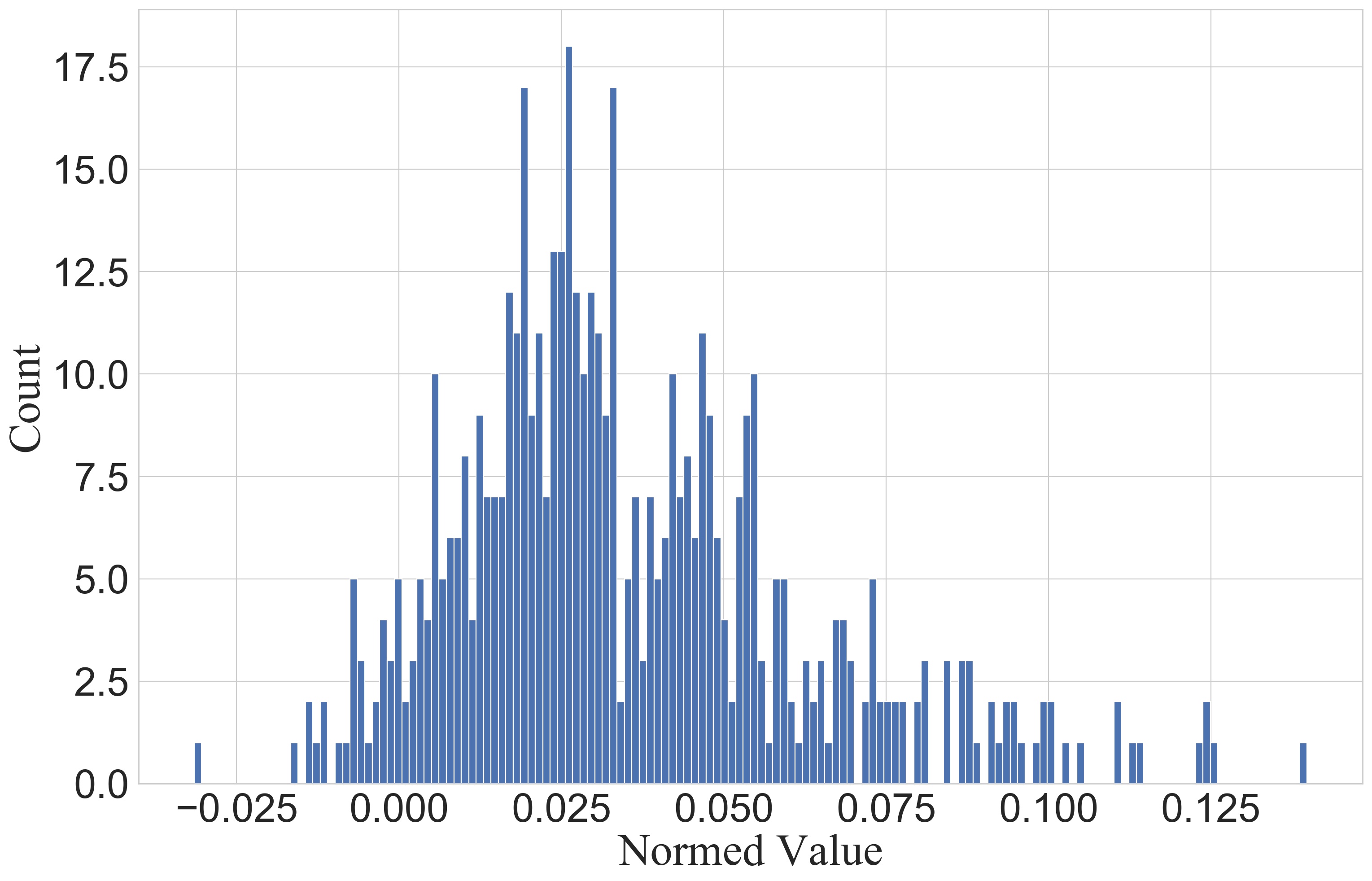}
}
\quad
\subfigure[GFBS distribution in $c$-8.]{
\includegraphics[width=3.4cm]{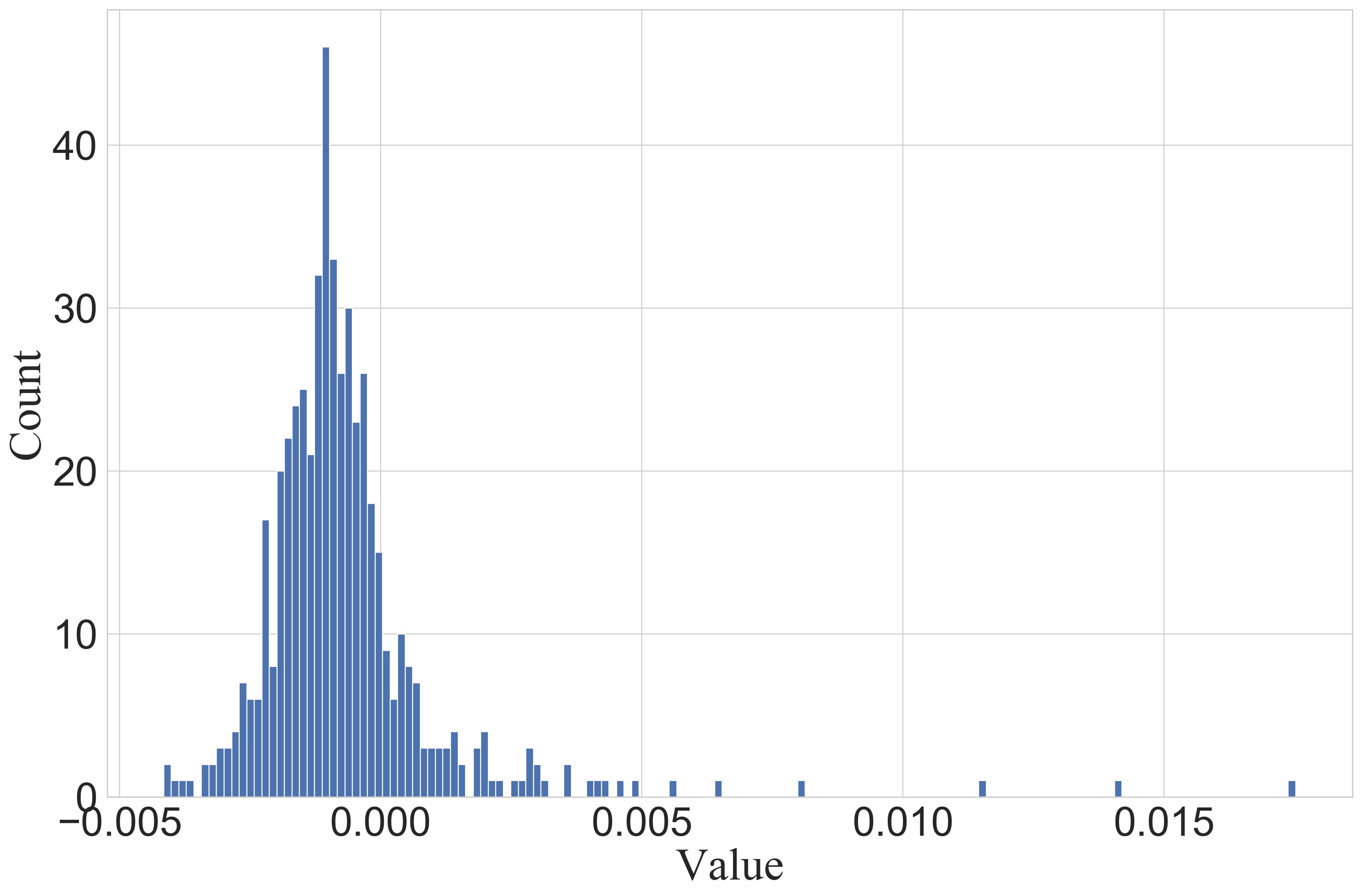}
}
\quad
\subfigure[GFBS distribution in $c$-9.]{
\includegraphics[width=3.4cm]{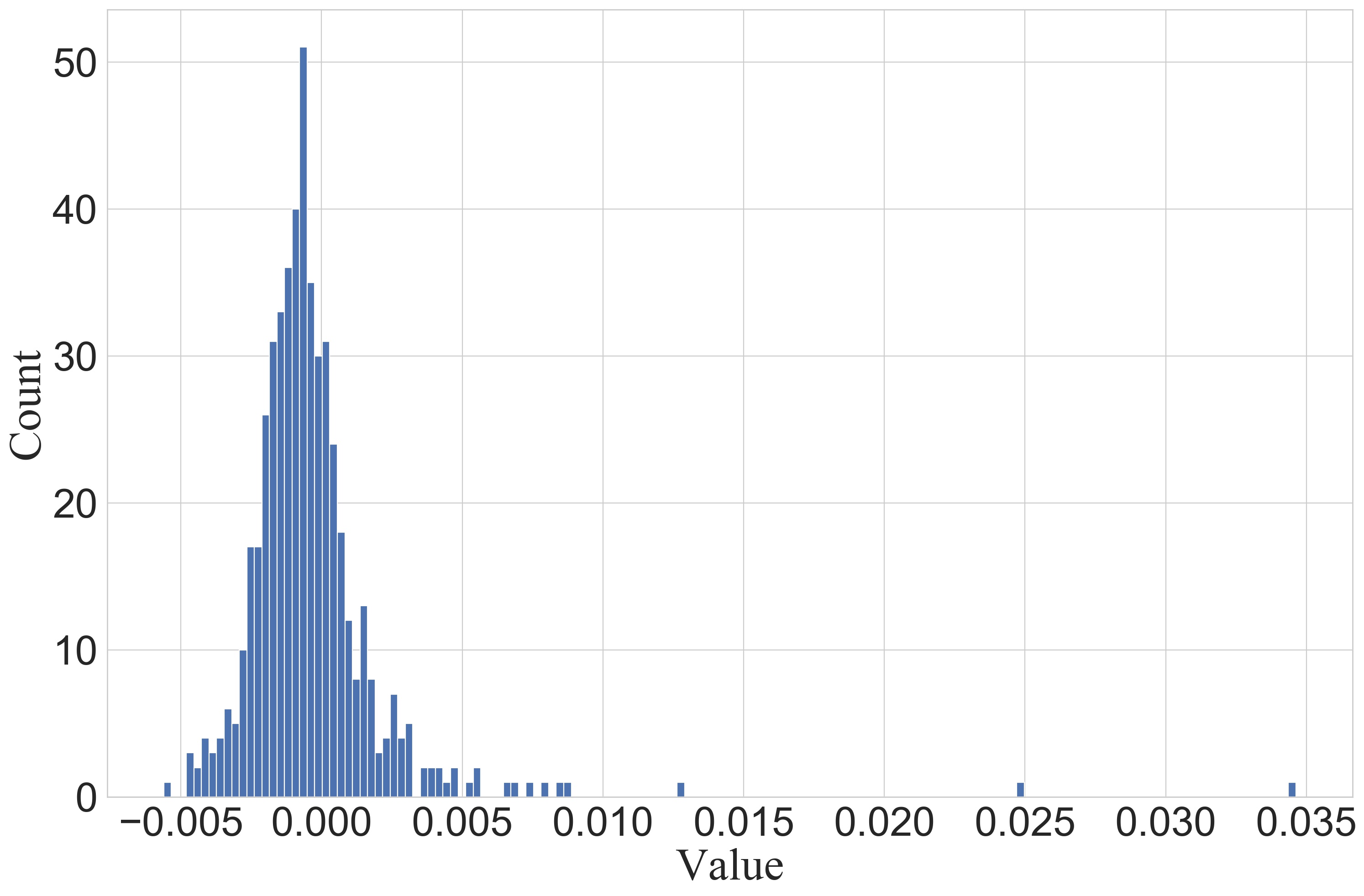}
}
\quad
\subfigure[GFBS distribution in $c$-10.]{
\includegraphics[width=3.4cm]{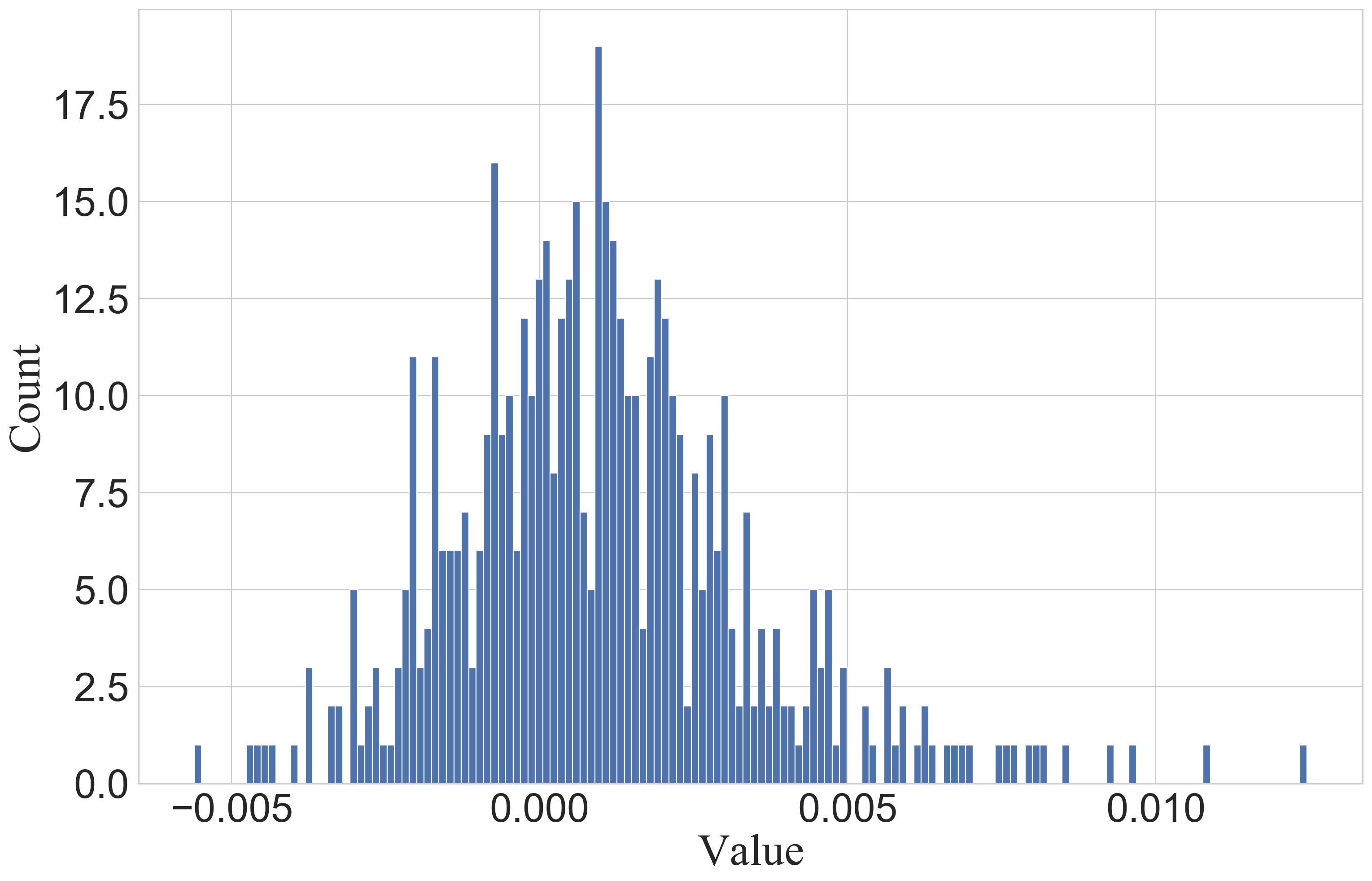}
}
\quad
\subfigure[GFBS distribution in $c$-11.]{
\includegraphics[width=3.4cm]{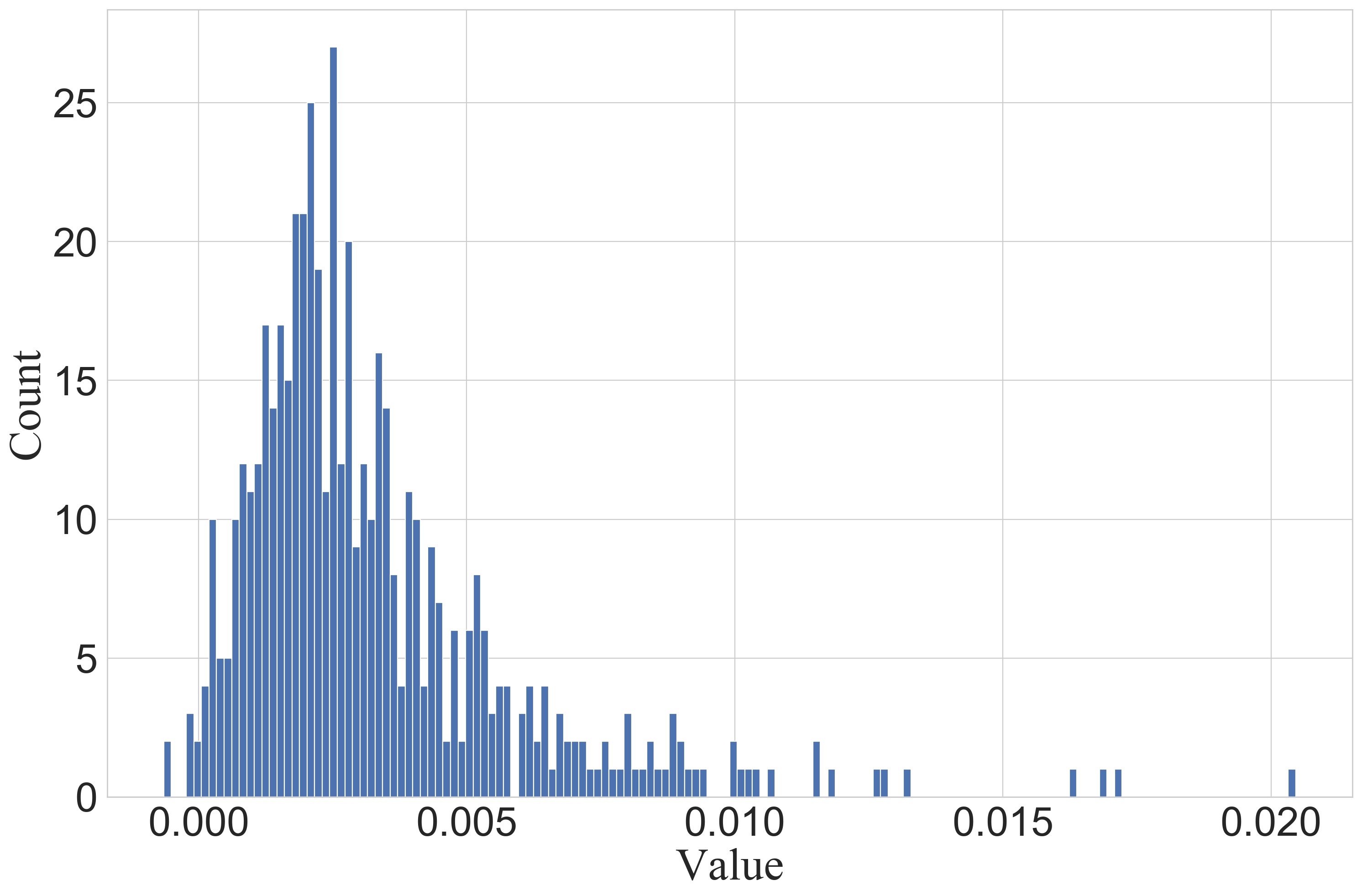}
}
\vspace{-12pt}
\caption{The distribution of the normed scaling parameter $\gamma$, normed shifting parameter $\beta$, and the gradient based channel saliency (GFBS) proposed in this paper in layer 8, 9, 10, 11 of the VGGNet-16. $c$-$i$ denotes the $i$-th convolutional layer.}
\vspace{-12pt}
\label{fig:distribution}
\end{figure*}
\subsection{Channel Saliencies Sorting and Pruning}
In this subsection, we introduce the pruning strategy based on the aforementioned theorem. Starting from a pre-trained network, we select a minibatch of data and pass it into the network following the traditions of \cite{lee2018snip, lin2020hrank, li2020eagleeye}. After the forward process, the first derivative of the scaling parameters are calculated layer-wise and saved following the chain rule. The parameters in the network are not updated to avoid saliency  {oscillations.} To perform the pruning globally and treat the channels in different layers equally, layer-wise normalization are implemented for $J(\gamma_l^{(j)})$, $\gamma_l^{(j)}$, and $\beta_l^{(j)}$ to restrict the range within $[-1, 1]$. Specifically, we use $\ell_2$ norm for simplicity. Then, the gradient flow based saliency  of each channel is computed following Theorem 3.3 and sorted. Given an arbitrary desired pruned ratio or  {resource constraints}, the channels are pruned in sequence according to the GFBS method until the required condition is satisfied.

\subsection{Statistics Distribution Visualization }

To better explain and demonstrate how the proposed gradient flow reflects the channel saliencies inside a network, we collect statistical information of a pre-trained VGGNet-16 network and visualize the distribution of different components in the network in Figure \ref{fig:distribution}. We show layer 8, 9, 10, 11 here because they have the same number of channels and have more samples than shallow layers. For the $\ell_2$-normed scaling parameter $\gamma$, all layers display a resembled distribution, demonstrating that the pruned ratio for each layer can be similar if only $\gamma$ is used as the criterion. The layer-wise distribution of $\beta$ appears to be entirely different. This phenomenon indicates that the activated proportion is varied regardless of the similarity of the feature distribution before ReLU. In the last row of the figure, \bp{GFBS} method is visualized and those channels with less importance can be easily identified and removed based on the value. It is also deduced that 
\bp{GFBS} method  {sorts} channel saliencies in a holistic perspective. For example, $c$-8 has more negative values than $c$-11 and $c$-9, $c$-10 both have a commensurate distribution. Therefore, more channels will be pruned in $c$-8 than $c$-11 instead of each layer is pruned in an approximate equal ratio.

\section{Experiments}

\subsection{Datasets and Implementation Details}
\label{4.1}
\textbf{Datasets.} To demonstrate the effectiveness of GFBS, we evaluate it on three distinct image classification datasets CIFAR-10 \cite{krizhevsky2009learning}, CIFAR100 \cite{krizhevsky2009learning} and ImageNet \cite{russakovsky2015imagenet} and image denoising benchmarks BSD68 and Set12. Due to space limitations, the results on CIFAR-100 are shown in the supplementary file. 
We compress representative DCNNs for evaluating the proposed method. On CIFAR-10, we use VGGNet-16 with BN layers \cite{simonyan2014very, ioffe2015batch}, ResNets \cite{he2016deep}, MobileNetv2 \cite{sandler2018mobilenetv2}, and DenseNet-40 \cite{huang2017densely}. For ImageNet, we prune ResNets with the proposed method. For the image denoising datasets, the representative network DnCNN \cite{zhang2017beyond} is utilized for pruning.\\
\textbf{Implementation Details.} Experiments are implemented with PyTorch \cite{paszke2019pytorch} and PaddlePaddle. For the image classification experiments, the baseline models for CIFAR-10 are trained from scratch for 160 epochs using SGD, with starting learning rate 0.1 and multiply by 0.2 at epoch 80 and 120. The PyTorch pre-trained models are directly used as the baseline models on ImageNet. For the image denoising task, we set the noise level as 50 and train the baseline model with the 400 images following common practice as \cite{zhang2017beyond}. \textit{With the proposed method, we only perform one forward and backpropagation for a minibatch thus almost no extra computation time is required before finetuning the pruned model.} For the finetuning, we follow the iterative configurations in HRank \cite{lin2020hrank} in CIFAR-10 and directly prune all unimportant channels and finetune the network for 100 epochs on ImageNet similar to \cite{fpgm}. For the denosing experiments, we finetune the pruned network for 50 epochs using the Adam optimizer. The initial learning rate is $1e^{-4}$ and is divided by 10 at the 40th epoch. The proposed method merely involves one hyper-parameter $\lambda$ which is the balancing term. Practically, we set it to 0.05 in all experiments and we find it hardly deteriorates the performance. The analysis of this coefficient is provided in the ablation study.
\begin{table}[t]
\caption{Pruning results of VGGNet-16 on CIFAR-10.}
\label{cifar10-vgg16}
\vspace{-12pt}
\begin{center}
\begin{small}
\begin{sc}
\begin{tabular}{lcccccr}
\toprule
Network & Method &  $\downarrow$ FLOPs & $\downarrow$ Acc \\
\midrule
\multirow{12}{*}{VGGNet-16}  
     & SSS & 41\% & 0.94\% \\
     & GAL-0.1 & 45\% & 0.54\% \\
     & \textbf{Ours} &  \textbf{46\%} & \textbf{0.28\%} \\
     \cmidrule{2-4}
     & CP & 50\% & \textbf{0.32\%} \\
     & T$\&$L & 64\% & 1.90\% \\
     & HRank & 65\% & 1.62\% \\
     & \textbf{Ours} &  \textbf{68\%} & 0.70\% \\
     \cmidrule{2-4}
     & \textbf{Ours} &  \textbf{70\%} & 1.13\% \\
     & HRank & 77\% & 2.73\% \\
     & \textbf{Ours} &  \textbf{78\%} & \textbf{1.87\%} \\
     & T$\&$L & 81\% & 3.40\% \\
     & \textbf{Ours} &  \textbf{84\%} & \textbf{2.10\%} \\
\bottomrule
\end{tabular}
\end{sc}
\end{small}
\end{center}
\vspace{-18pt}
\end{table}

\begin{table}[t]
\caption{Pruning results of ResNets on CIFAR-10.}
\label{cifar10-resnets}
\vspace{-12pt}
\begin{center}
\begin{small}
\begin{sc}
\begin{tabular}{lcccccr}
\toprule
Depth & Method &  $\downarrow$ FLOPs & $\downarrow$ Acc \\
\midrule
\multirow{10}{*}{20}  
     & Variational & 16\% & 0.35\% \\
     & MorphNet & 25\% & 1.53\% \\
     & SSL & 26\% & 1.09\% \\
     & DSA & 26\% & 0.07\% \\
     & \textbf{Ours} & \textbf{27\%} & \textbf{-0.12\%} \\
     \cmidrule{2-4}
     & SFP &  40\% & 1.37\% \\
     & Rethink &  40\% & 1.34\% \\
     & FPGM & 42\% & 1.11\% \\
     & DSA & 50\% & 0.79\% \\
     & Hinge & \textbf{55\%} & 0.70\% \\
     & \textbf{Ours} &  53\% & \textbf{0.53\%} \\
\midrule
\multirow{6}{*}{32} 
     & MIL & 31\% & 1.59 \\
     & \textbf{Ours} & \textbf{33\%} & \textbf{-0.25} \\
     & SFP & 41\% & 0.55 \\
     & FPGM & 41\% & 0.32 \\
     & CNN-FCF & 42\% & 0.25 \\
     & \textbf{Ours} & \textbf{42\%} & \textbf{0.02} \\
\midrule
\multirow{12}{*}{56}
     & Variational & 20\% & 0.78\% \\
     & PNNREI & 28\% & 0.70\% \\
     & PFEC & 28\% & -0.02\% \\
     & LeGR & 30\% & -0.20\%\\
     & HRank & 30\% & -0.26\%\\
     & \textbf{Ours} & \textbf{42\%} & \textbf{-0.49\%} \\
     \cmidrule{2-4}
     & CP & 50\% & 1.00\% \\
     & AMC & 50\% & 0.90\% \\
     & Rethink & 50\% & 0.73\% \\
     & TAS & 53\% & 0.77\% \\
     & DAIS & 55\% & 0.82\% \\
     & \textbf{Ours} & \textbf{57\%} & \textbf{0.70\%} \\
\bottomrule
\end{tabular}
\end{sc}
\end{small}
\end{center}
\vspace{-12pt}
\end{table}

\begin{table}[t]
\caption{Pruning results of MobileNetv2 on CIFAR-10.}
\label{cifar10-mobilenetv2}
\vspace{-12pt}
\begin{center}
\begin{small}
\begin{sc}
\begin{tabular}{lcccccr}
\toprule
Network & Method &  $\downarrow$ FLOPs & $\downarrow$ Acc \\
\midrule
\multirow{6}{*}{MobileNetv2}  
     & WM & 26\% & 0.45\% \\
     & Rand DCP & 26\% & 0.57\% \\
     & DCP & 26\% & \textbf{-0.22\%} \\
     & \textbf{Ours} &  \textbf{31\%} & -0.09\% \\
     \cmidrule{2-4}
     & SCOP & 40\% & 0.24\% \\ 
     & \textbf{Ours} &  \textbf{42\%} & \textbf{0.23\%} \\
\bottomrule
\end{tabular}
\end{sc}
\end{small}
\end{center}
\vspace{-12pt}
\end{table}
\subsection{Experiments on Image Classification}
\subsubsection{Results on CIFAR-10.} To validate the efficacy of the proposed method, we first conduct experiments on the CIFAR-10 dataset to discover the saved computation consumption ratio and the final accuracy of the sub-networks generated with our method. Four prototypical but distinct architectures are considered, which are VGGNet-16, ResNet series, MobileNetv2, and DenseNet-40. We present the percentage of pruned FLOPs and the accuracy drop against the baseline models. 

\noindent \textbf{VGGNet-16.} We use the VGGNet-16 equipped with the BN after each convolutional layer which is widely recognized today and display the results in Table \ref{cifar10-vgg16}. The results of GAL \cite{gal}, T\&L \cite{tandl}, and HRank \cite{lin2020hrank} are from their original paper. SSS \cite{sss} is based on the reproduction by GAL. Results of CP \cite{cp} is collected from the reproduction on CIFAR-10. Compared with SSS, GAL, and CP, our method shows a notable trade-off between the pruned FLOPs and the accuracy of the generated sub-network (46\% pruned FLOPs with only 0.28\% accuracy drop). When comparing with HRank and T\&L, our method outperforms them under more strict resource constraints. Specially, GFBS method is able to prune up to 84\% FLOPs and still have a slight accuracy drop of 2.10\% meanwhile HRank and T\&L prune 77\% and 81\% FLOPs but have a decline in accuracy of 2.73\% and 3.40\%.

\noindent \textbf{ResNets.} We test the performance of GFBS on ResNet-20, 32, and 56. Results are given in Table \ref{cifar10-resnets}. Among which, results of Variational \cite{zhao2019variational}, DSA \cite{ning2020dsa}, SFP \cite{SFP}, Rethink \cite{liu2018rethinking}, FPGM \cite{fpgm}, Hinge \cite{hinge}, MIL \cite{mil}, CNN-FCF \cite{cnn-fcf}, PFEC \cite{li2016pruning}, LeGR \cite{legr}, HRank \cite{lin2020hrank}, CP \cite{cp}, AMC \cite{amc}, TAS \cite{tas}, DAIS \cite{dais} are directly taken from their official papers. Results of MorphNet \cite{gordon2018morphnet} and SSL \cite{ssl} are from the implementation by DSA and result of PNNREI \cite{molchanov2016pruning} is provided in LeGR. It is observed that the proposed method shows a competitive performance in pruning ResNet series. For ResNet-20 and 32, our method can improve the base accuracy by 0.12\% and 0.25\% with $\sim$30\% FLOPs reduction, which surpasses other approaches by a large margin. When comparing the performance in ResNet-56, the proposed GFBS method consistently outperforms other methods in both FLOPs and accuracy. Concretely, the GFBS pruned sub-network with 57\% FLOPs reduction only shows a 0.70\% accuracy drop, which is still better than \cite{zhao2019variational} that prune 20\% FLOPs (0.78\% accuracy drop). Moreover, our method also outperforms other methods that only saved 50\% of the computation resource.

\noindent \textbf{MobileNetv2.} The experiment results of the MobileNetv2 are reported in Table \ref{cifar10-mobilenetv2}. WM, Rand DCP, and DCP are collected from DCP paper \cite{dcp} and SCOP \cite{tang2020scop} is from their original paper. Because MobileNetv2 is already a compact architecture, it is more challenging to get pruned. However, our method can still obtain 42\% reduction rate in FLOPs meanwhile only has a degradation of 0.23\% in accuracy, demonstrating that the proposed GFBS is evidently superior than the previous state-of-the-art approaches.

\noindent \textbf{DenseNet40.} Table \ref{cifar10-densenet} presents the results for DenseNet-40. The results of other counterparts are collected from their papers \cite{zhao2019variational, slimming, lin2020hrank, kang2020operation}. Generally, our method shows a far better test accuracy when pruning same FLOPs over other methods (Variational, SCP). This phenomenon indicates that the proposed method not only is able to implement on arbitrary moderns DCNNs, but also shows a more efficient performance. Specifically for the dense connections, GFBS computes the saliencies for each concatenation with respect to their embedded information in both BN and ReLU thus can identify the unimportant channels more accurately. \textit{The superiority in performance also validates our conjecture mentioned in Subsection 3.1 that previous methods that only consider BN may not provide a holistic enough evaluation of channel saliency.}

\begin{table}[t]
\caption{Pruning results of DenseNet-40 on CIFAR-10.}
\label{cifar10-densenet}
\vspace{-12pt}
\begin{center}
\begin{small}
\begin{sc}
\begin{tabular}{lcccccr}
\toprule
Network & Method &  $\downarrow$ FLOPs & $\downarrow$ Acc \\
\midrule
\multirow{6}{*}{DenseNet-40}  
     & Variational & 45\% & 0.95\% \\
     & \textbf{Ours} &  \textbf{45\%} & \textbf{-0.23\%} \\
     \cmidrule{2-4}
     & Slimming & 55\% & 0.46\% \\
     & HRank & 61\% & 1.13\% \\ 
     & SCP & 71\% & 0.62\% \\
     & \textbf{Ours} &  \textbf{71\%} & \textbf{0.41\%} \\
\bottomrule
\end{tabular}
\end{sc}
\end{small}
\end{center}
\vspace{-18pt}
\end{table}

\subsubsection{Results on ImageNet.} To validate the effectiveness of the proposed method on large-scale image classification datasets, we further apply GFBS on the ImageNet dataset and compare with more methods including \cite{luo2017thinet, gdp, liebenwein2019provable}. We prune ResNet-18 and ResNet-50 and provide the results in Table \ref{imagenet}. It can be observed that the proposed method outperforms other channel pruning methods in terms of both FLOPs reduction ratio and accuracy drop, demonstrating that the proposed method is also capable of recognizing channel saliencies better for large-scale datasets.

\begin{table}[b]
\caption{Pruning results of ResNets on ImageNet.}
\label{imagenet}
\vspace{-12pt}
\begin{center}
\begin{small}
\begin{sc}
\begin{tabular}{lcccccr}
\toprule
Depth & Method &  $\downarrow$ FLOPs & $\downarrow$ Acc1 & $\downarrow$ Acc5 \\
\midrule
\multirow{5}{*}{18}  
     & MIL & 33\% & 3.65\% &2.30\%\\
     & SFP & 42\% & 3.18\% &1.85\%\\
     & FPGM & 42\% & 1.87\% &1.15\%\\
     & PFP-B & 43\% & 4.09\% &2.32\%\\
     & \textbf{Ours} &  \textbf{43\%} & \textbf{1.83\%} &\textbf{1.06\%}\\
\midrule
\multirow{5}{*}{50} 
     & CP & 34\% & 3.85\% & 2.07\% \\
     & ThiNet & 37\% & 4.11\% & -- \\
     & GDP & 41\% & 2.52\% & 1.25\% \\
     & GAL-0.5 & \textbf{42\%} & 4.20\% & 1.93\% \\
     & \textbf{Ours} & 37\% & \textbf{2.27\%} & \textbf{1.04\%} \\
\bottomrule
\end{tabular}
\end{sc}
\end{small}
\end{center}
\vspace{-16pt}
\end{table}

\begin{figure*}[t]
\centering
\subfigure[VGGNet-16.]{
\includegraphics[width=3.9cm]{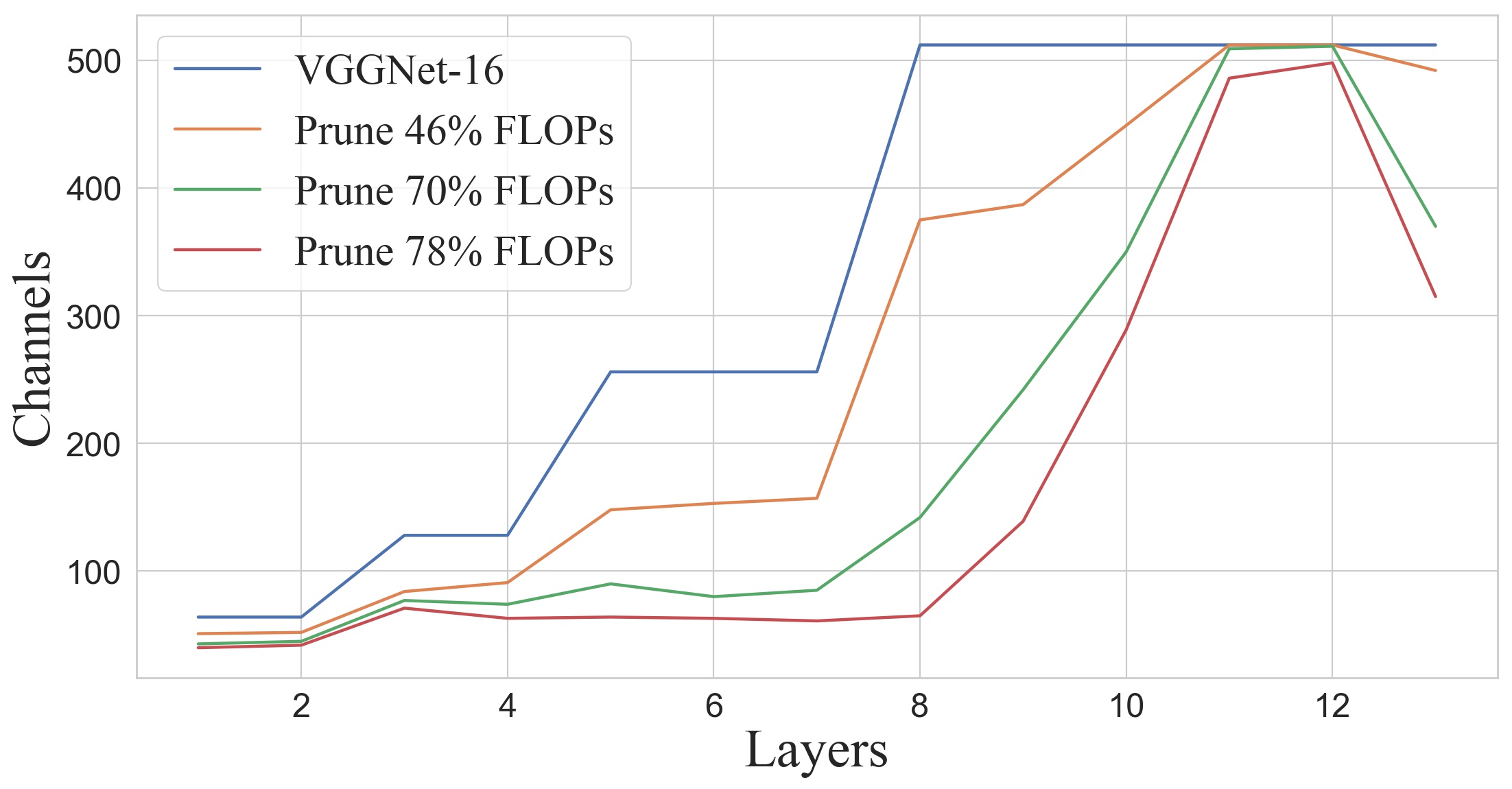}
}
\quad
\subfigure[ResNet-20.]{
\includegraphics[width=3.9cm]{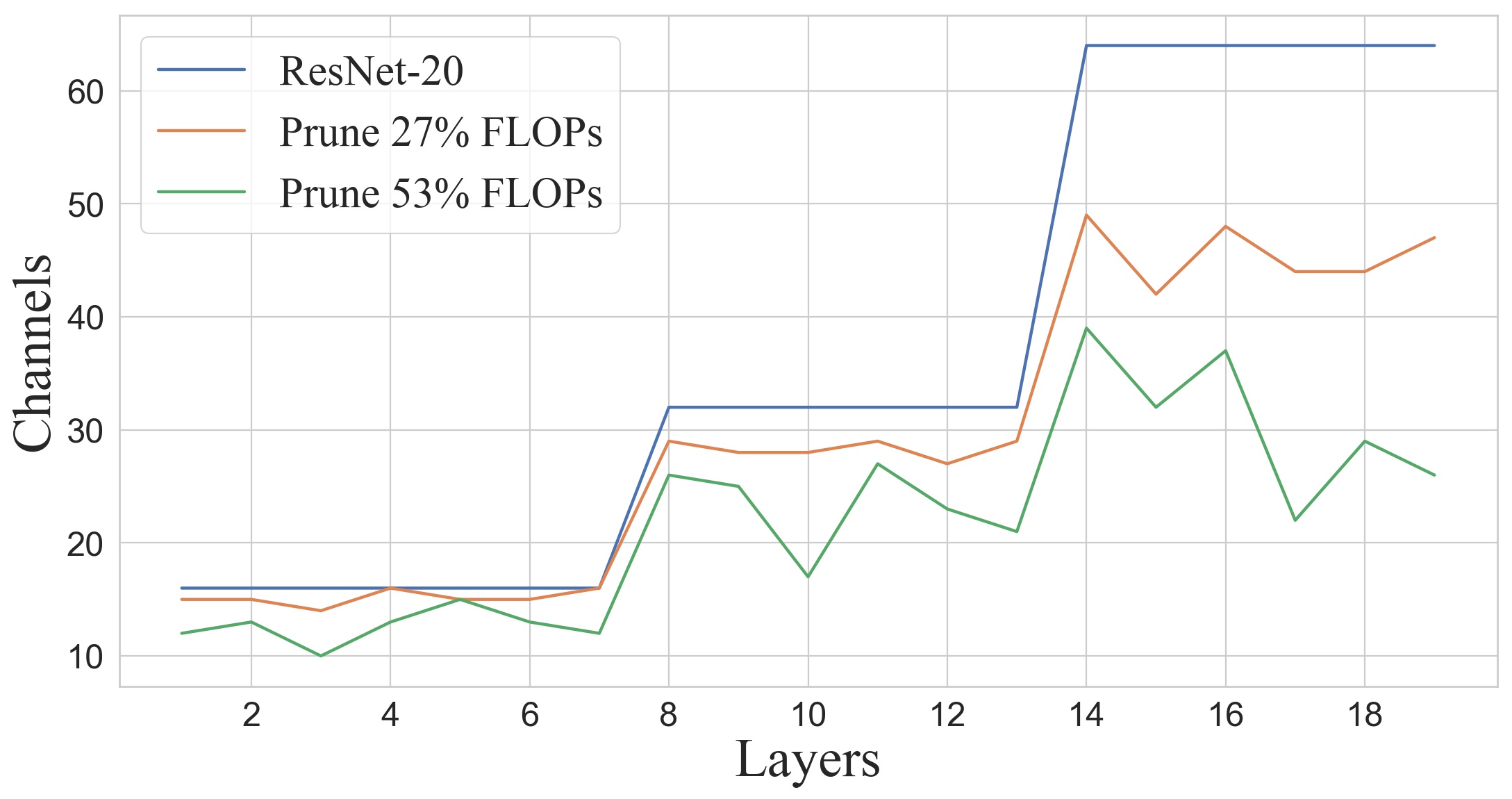}
}
\quad
\subfigure[ResNet-32.]{
\includegraphics[width=3.9cm]{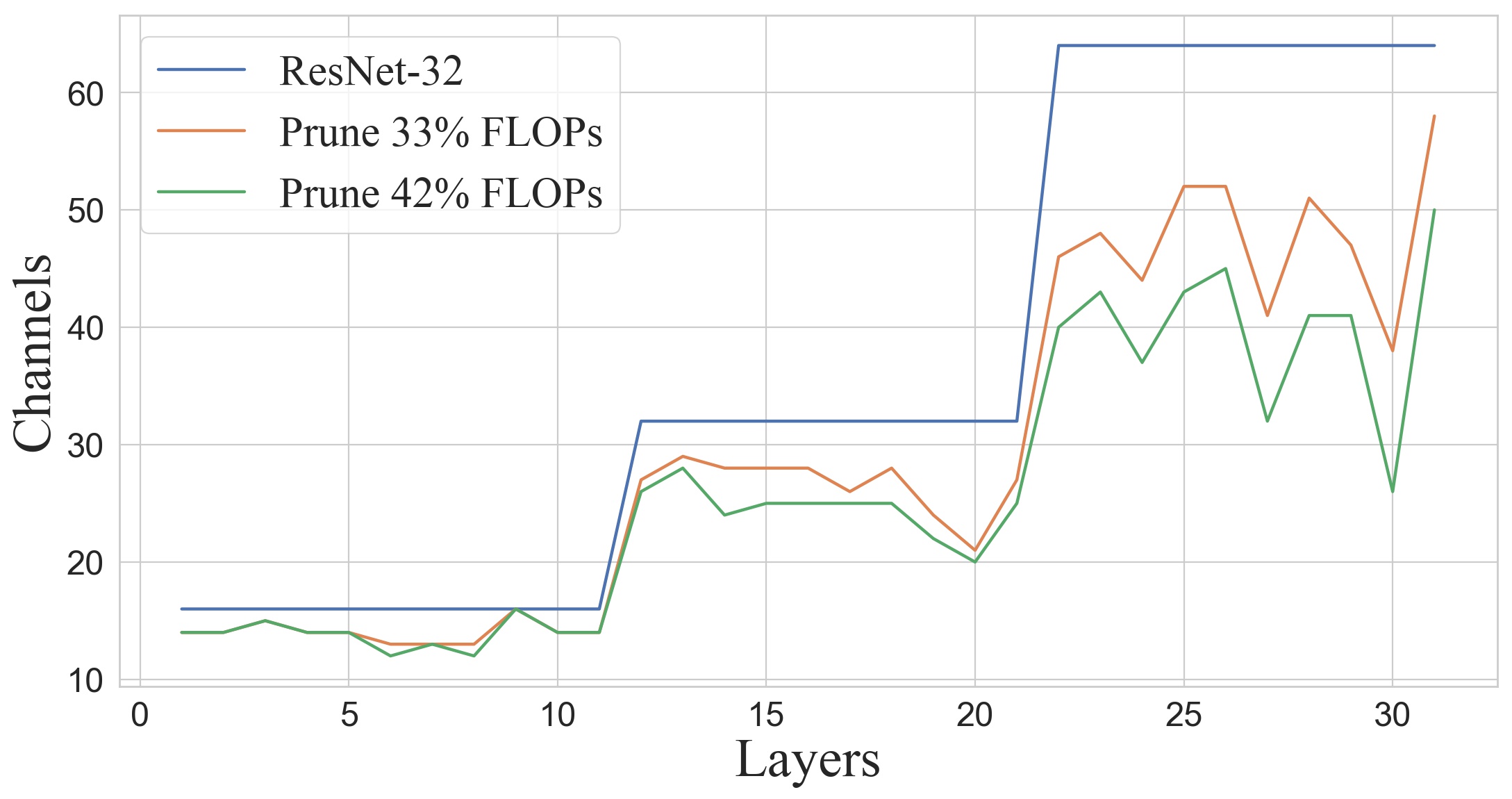}
}
\quad
\subfigure[ResNet-56.]{
\includegraphics[width=3.9cm]{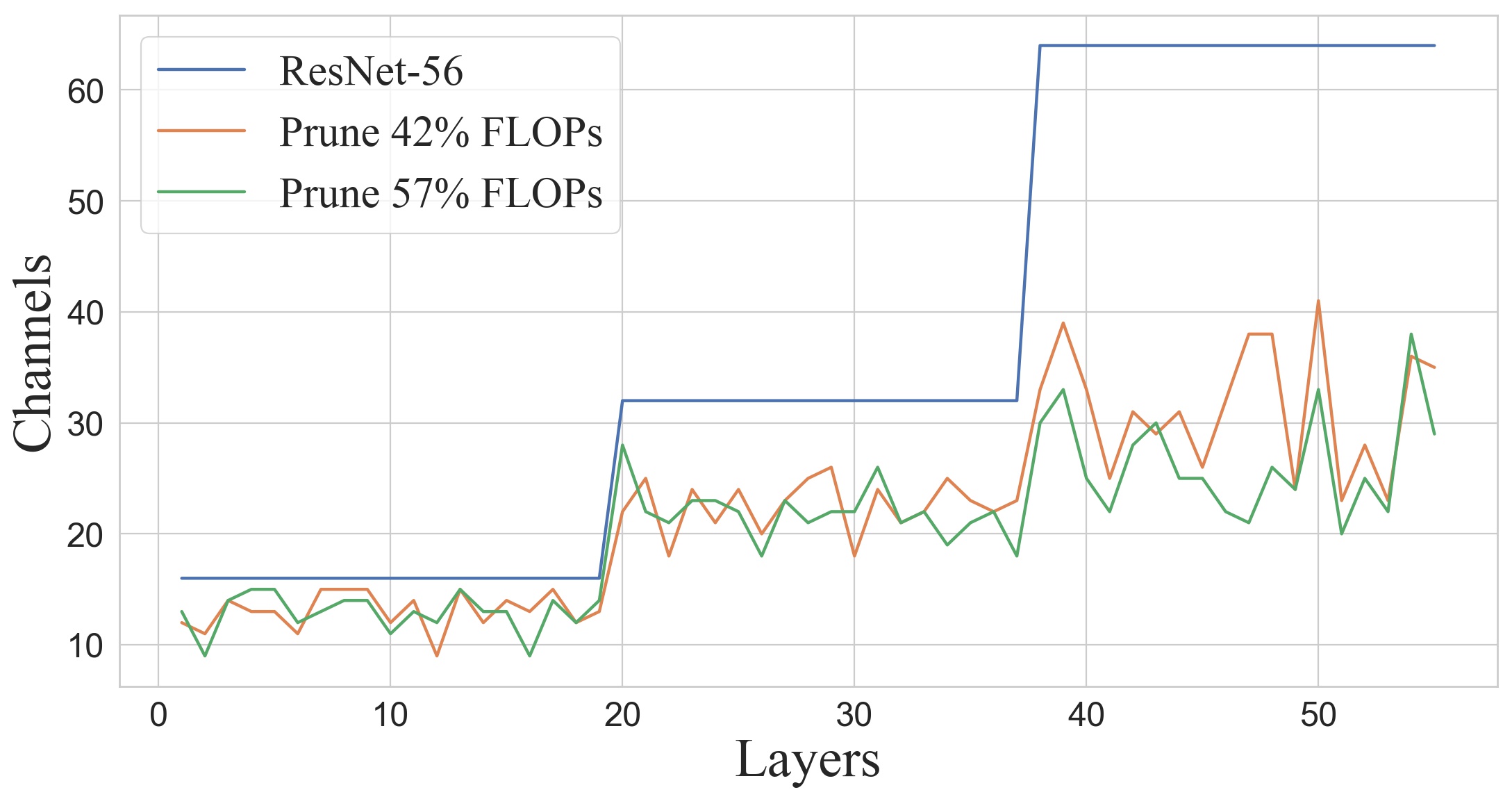}
}
\vspace{-12pt}
\caption{The sub-networks generated with the proposed method. From left to right: (a) VGGNet-16, (b) ResNet-20, (c) ResNet-32, and (d) ResNet-56. Each solid line refers to a sub-network with a certain FLOPs constraint.}
\label{fig:remaining_channels}
\vspace{-12pt}
\end{figure*}

\begin{figure}[h]
    \centering
    
    \includegraphics[width=0.40\textwidth]{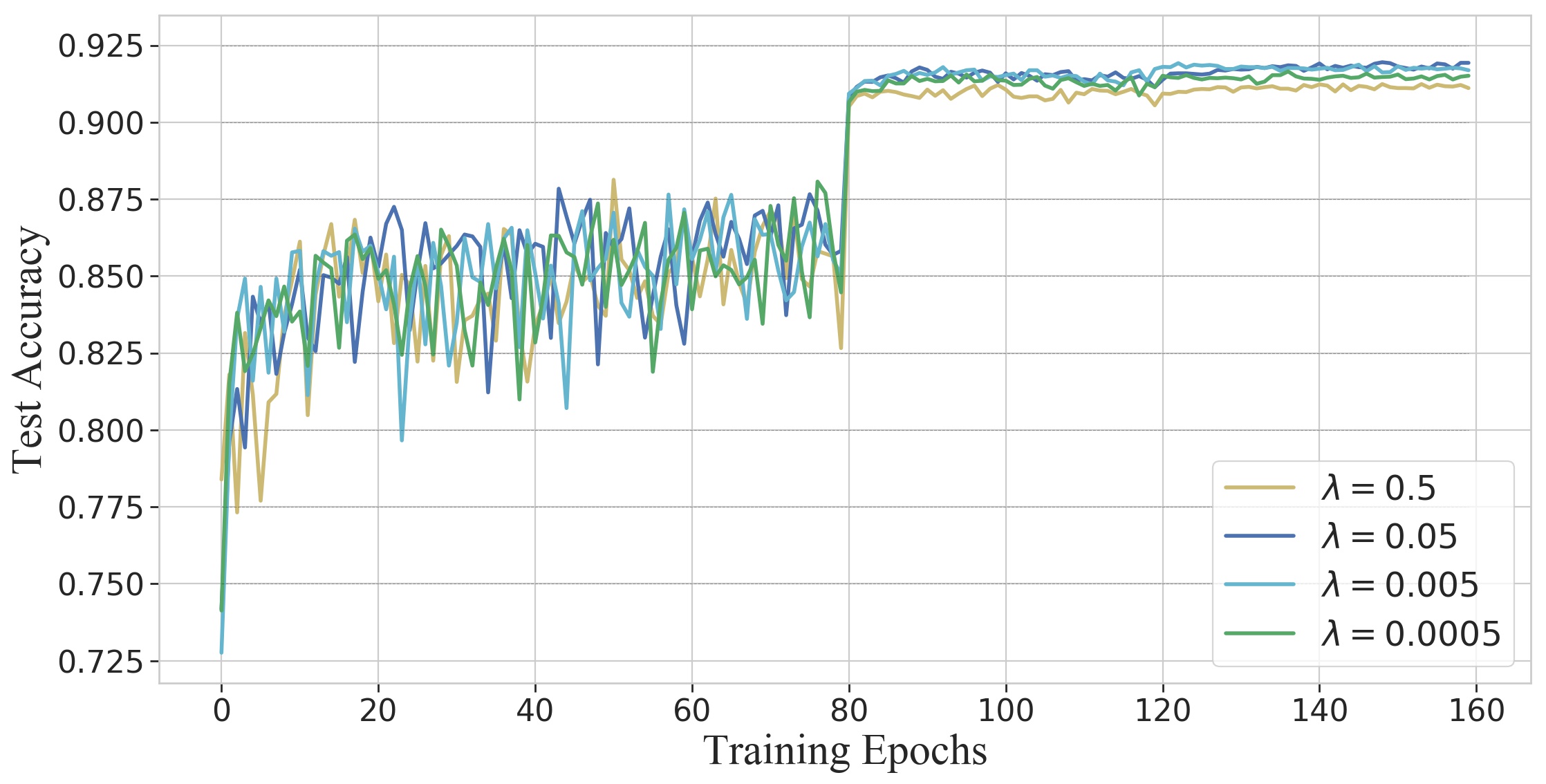}
    \vspace{-12pt}
    \caption{Analysis of the balancing term $\lambda$ on ResNet-20.}
    \label{fig:lambda}
    \vspace{-6pt}
\end{figure}

\begin{figure}[h]
    \centering
    
    \includegraphics[width=0.48\textwidth]{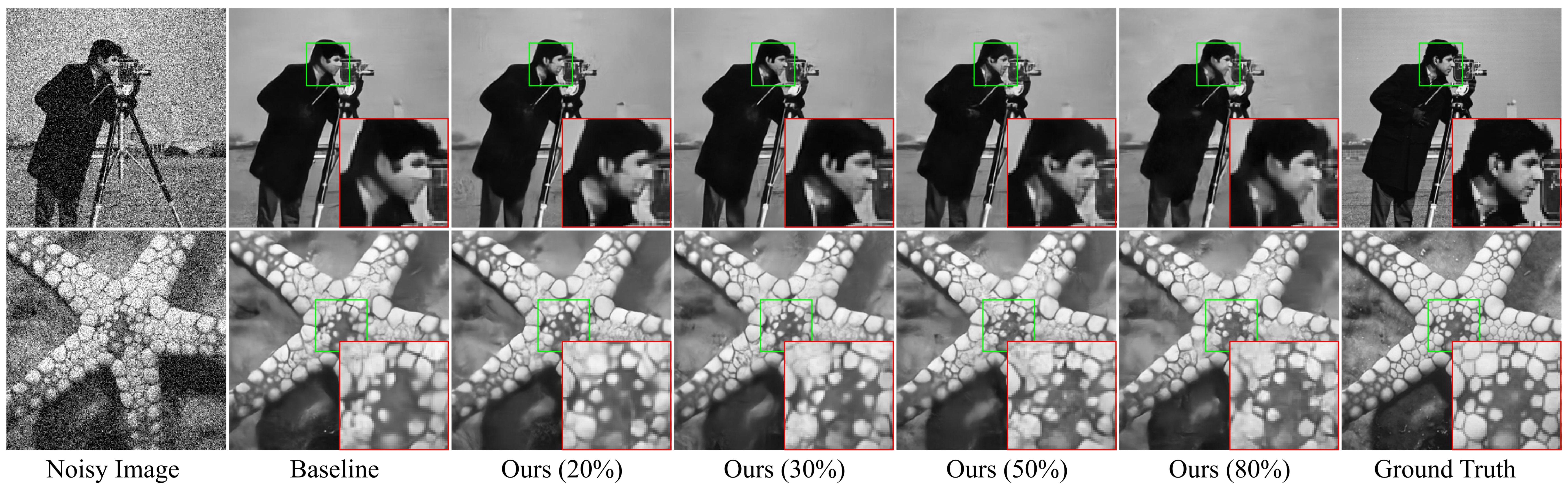}
    \vspace{-19pt}
    \caption{Visualized denoising results with noise level 50.}
    \label{fig:denoise}
    \vspace{-10pt}
\end{figure}

\subsection{Experiments on Image Denoising}
To validate the efficiency of the proposed method on different multimedia tasks, experiments on gray image denoising are conducted and the results are recorded in Table \ref{denoise}. The baseline network is \bp{DnCNN ~\cite{zhang2017beyond} that is popular in the field of image denoising. Four different pruning strategies including prune 20\%, 30\%, 50\%, and 80\% channels are tested. It should be noted that since most previous model pruning methods do not test their schemes on image denosing task, we only re-implement \cite{slimming} and compare with it. As demonstrated in the Table, pruning 30\% and 46\% of the FLOPs with the proposed method have a negligible affect on the PSNR, with only 0.01dB and 0.02dB reduction on the BSD68 dataset and 0.01dB and 0.04dB reduction on the Set12 dataset, respectively. When pruning the FLOPs to a very large extent of 71\%, the algorithm's denoising performance in terms of PSNR on BSD68 and Set12 only decrease 0.05dB and 0.07dB respectively, showing the efficacy of the proposed GFBS for the image denoising task. We further significantly cut down the FLOPs to 94\%, PSNR decrease of 0.28dB and 0.45dB on those two datasets are observed, illustrating that using the proposed method, a favorable performance of image denoising can still be obtained even with a large model compression ratio. The visualized denoising results are shown in Figure \ref{fig:denoise}. It may be noticed that the denoised images still show promising results when the model is pruned to different degrees.}

\begin{table}[t]
\caption{Pruning results of DnCNN on BSD68 and Set12.}
\label{denoise}
\vspace{-12pt}
\begin{center}
\begin{small}
\begin{sc}
\begin{tabular}{lcccc}
\toprule
\multirow{2}{*}{Network} & \multirow{2}{*}{Method} & \multirow{2}{*}{$\downarrow$ FLOPs} & \multicolumn{2}{c}{PSNR (dB)} \\\cline{4-5} 
                         &                         &      & BSD68       & Set12      \\
\midrule
\multirow{9}{*}{DnCNN}   & Baseline                & --   & 26.22       & 27.16      \\
                         & Slimming (20\%)             & 30\% & 26.20 & 27.11      \\
                         & Ours (20\%)             & 30\% & 26.21       & 27.15      \\
                         & Slimming (30\%)             & 45\% & 26.20       & 27.12      \\
                         & Ours (30\%)             & 46\% & 26.20       & 27.12      \\
                         & Slimming (50\%)             & 70\% & 26.17       & 27.04      \\
                         & Ours (50\%)             & 71\% & 26.17       & 27.09      \\
                         & Slimming (80\%)             & 92\% & 25.87       & 26.49    \\
                         & Ours (80\%)             & 94\% & 25.94       & 26.71      \\
\bottomrule
\end{tabular}
\end{sc}
\end{small}
\end{center}
\vspace{-8pt}
\end{table}

\subsection{Ablation Studies}

\textbf{Components in the Proposed Model Pruning Scheme.} The proposed GFBS is a combination of the first-order Taylor polynomial of $\gamma$ and the signed $\beta$ in the BN layer. To demonstrate the validity of this assemble, ablation study on the two components are conducted. Following the implementation in Subsection \ref{4.1}, we test the performance of GFBS with the first part only (GFBS-$\gamma$), GFBS with the second part only (GFBS-$\beta$), and the complete GFBS. For all methods, we select channels with the bottom 20\% saliencies in ResNet-20 with regard to the criteria. Results in Table \ref{ablationcomponents} show that using the components in GFBS individually will harm the performance while the combination can discover the channels with the lowest saliencies and improve the accuracy after removing them. 
\begin{table}
\caption{Ablation study of the components.}
\label{ablationcomponents}
\vspace{-10pt}
\begin{center}
\begin{small}
\begin{sc}
\begin{tabular}{lcccccr}
\toprule
  Network & $\downarrow$ Channels & Criterion & $\downarrow$ Acc  \\
\midrule
\multirow{3}{*}{ResNet-20}&\multirow{3}{*}{20\%} & GFBS-$\gamma$ & 0.04\%  \\
                          &                      & GFBS-$\beta$ & 0.44\%  \\
                          &                      & \textbf{GFBS} & \textbf{-0.12\%}  \\
\bottomrule
\end{tabular}
\end{sc}
\end{small}
\end{center}
\vspace{-12pt}
\end{table}

\noindent \textbf{Analysis of the Pruned Channels.} The configurations of the remaining channels for each layer in the pruned VGGNet-16 and ResNets are plotted in Figure \ref{fig:remaining_channels}. We may observe the following phenomenons. \textit{The channel saliencies are uniformly distributed among layers.} The discovered sub-networks have similar structures as the unpruned networks, which means: 1) The channel saliencies sorting is conducted in a holistic way instead of layer-by-layer, thus the proposed method hardly suffers from layer collapse, {which means that all channels in a layer are pruned.} 2) The proposed method may not damage the original design but only sieving the redundant channels to form a more compact network. \textit{Shallow layers are more likely to be pruned than deep ones.} It is noticed that deep layers tend to have higher tendencies to get preserved since they have higher saliencies. These layers contain more high-level semantic information than shallow layers thus are more crucial.
A similar phenomenon is also observed in other works \cite{zhao2019variational, liu2019metapruning}.

\noindent \textbf{Impact of $\lambda$.} We have proven that the authentic channel saliencies should be the combination of 
$J(\gamma_l^{(j)}) \gamma_l^{(j)}$ and $\beta_l^{(j)}$.
However, we have performed layer-wise normalization for these two parts and a coefficient for balancing needs to be determined. To select an appropriate hyper-parameter, we evaluate the performance of the pruned ResNet-20 on CIFAR-10 dataset with $\lambda$ values across four magnitudes. The pruned network is finetuned for 160 epochs with learning rate decay at the 80 and 120 epoch. Results in Figure \ref{fig:lambda} have shown that $\lambda=0.05$ outperforms the other settings and achieves the best trade-off between the contribution of the two components in the proposed gradient flow based saliency method.

\vspace{-5pt}
\section{Conclusions}

In this work, we have proposed a novel structured pruning approach via gradient flow based saliency in the deep neural networks named GFBS. The proposed new method utilizes Taylor expansion in a holistic perspective to approximate oracle pruning, and the channel saliencies can be represented by the weighted combination of the first-order of the scaling parameters and the signed shifting parameters in BN via deduction. 
Compared with previous methods, the proposed scheme not only considers the impact of conv layers but also BN and ReLU layers. It prunes the network globally, which can be implemented within one run of the network and does not rely on the magnitude of the filters. Besides, it demands no trainable parameters and is easy to plug in any modern architecture. 
{Extensive} experimental results on {both image classification and image denoising} have demonstrated the efficacy and superiorty of the proposed channel pruning method.

\bibliographystyle{ACM-Reference-Format}
\balance
\bibliography{sample-base}


\end{document}